%% file: main.tex
\documentclass[11pt, a4paper]{article}
\usepackage[left=1in, right=1in, top=1in, bottom=1in]{geometry}

\usepackage[utf8]{inputenc}
\usepackage{lmodern}
\usepackage[T1]{fontenc}
\usepackage[english]{babel}
\usepackage{ifpdf}

\usepackage[dvipsnames]{xcolor}
\usepackage[colorlinks=true,linkcolor=teal,citecolor=teal]{hyperref}

\hypersetup{
	pdftitle={},
	pdfauthor={}
} 

\usepackage{graphicx, amsmath, amsthm, amssymb, enumitem, mathrsfs} 
\usepackage{subcaption}
\usepackage{booktabs} 

\usepackage[capitalize]{cleveref}
\usepackage[square,numbers]{natbib}
\usepackage{soul}

\usepackage{tikz}
\usepackage{tikz-3dplot} 
\usetikzlibrary{3d} 
\usetikzlibrary{arrows.meta}
\usepackage{filecontents}
\usepackage{pgfplots, pgfplotstable}
\usepgfplotslibrary{statistics}
\usepackage{caption}

\newcommand{\pr}[1]{\mathbb{P}\left({#1}\right)}
\newcommand{\ep}[1]{\mathbb{E}\left({#1}\right)}
\newcommand{\va}[1]{\mathrm{Var}\left({#1}\right)}
\newcommand{\ol}[1]{\overline{#1}}

\newcommand{\op}[1]{\left\|{#1}\right\|_{\mathrm{op}}}
\newcommand{\fro}[1]{\left\|{#1}\right\|_{\mathrm{F}}}
\newcommand{\two}[1]{\left\|{#1}\right\|_{2}}
\newcommand{\card}[1]{\#{#1}}

\newcommand{\eprob}{p}
\newcommand{\sprob}{s}
\newcommand{\signature}{\sigma}
\newcommand{\frustration}{\eta}
\newcommand{\eprmatrix}{\mathbf{P}}
\newcommand{\sprmatrix}{\mathbf{S}} 
\newcommand{\signedblock}{\mathsf{SignedBlock}}

\newtheorem{theorem}{Theorem}[section]
\newtheorem{corollary}[theorem]{Corollary}
\newtheorem{remark}[theorem]{Remark}
\newtheorem{definition}[theorem]{Definition}
\newtheorem{lemma}[theorem]{Lemma}
\newtheorem{proposition}[theorem]{Proposition}

\usepackage{xcolor, soul}

\title{Matrix Concentration for Random Signed Graphs and Community Recovery in the Signed Stochastic Block Model}
\author{Sawyer Jack Robertson}
\date{}

\begin{document}

    \captionsetup[figure]{labelfont={bf},name={Figure},labelsep=period}
	\maketitle

    \vspace*{-0.375in}
    \begin{abstract}
        We consider graphs where edges and their signs are added independently at random from among all pairs of nodes. We establish strong concentration inequalities for adjacency and Laplacian matrices obtained from this family of random graph models. Then, we apply our results to study graphs sampled from the signed stochastic block model. Namely, we take a two-community setting where edges within the communities have positive signs and edges between the communities have negative signs and apply a random sign perturbation with probability $0< s <1/2$. In this setting, our findings include: first, the spectral gap of the corresponding signed Laplacian matrix concentrates near $2s$ with high probability; and second, the sign of the first eigenvector of the Laplacian matrix defines a weakly consistent estimator for the balanced community detection problem, or equivalently, the $\pm 1$ synchronization problem. We supplement our theoretical contributions with experimental data obtained from the models under consideration.

        \vspace{.1in}
        {\footnotesize\hspace*{-.2in}\textbf{\texttt{MSC2020:}} 62H30, 05C80, 05C22

        \hspace*{-.2in}\textbf{\texttt{Keywords:}} signed graphs, graph Laplacians, matrix concentration, stochastic block models, community detection, synchronization}
    \end{abstract}


    \section{Introduction}

    The study of signed graphs, i.e., graphs whose edges carry positive or negative labels, dates back at least sixty-five years to the work of Cartwright and Harary~\cite{cartwright1956structural}. In many real-world networks, relationships are not merely present or absent, but can be cooperative or antagonistic. For example, in social networks, edges may represent friendships or enmities; in biological contexts, they may encode synergistic or inhibitive interactions. Therefore it can be said that signed graphs provide a more nuanced modeling tool compared to traditional graphs, allowing researchers to study phenomena such as community structure, conflict, and alignment under one unifying framework. As such, random signed graphs can be used to model and understand various techniques from the world of data science that draw on sign-based structures.
    
    In this paper we consider a general class of random signed graphs constructed as follows. Let $n\geq 2$ be a fixed integer and let $[n] = \{1, 2, \dotsc, n\}$ be a set of nodes. For each pair $e = \{i, j\}\in {[n]\choose 2}$, add $e$ to the edge set of the graph with probability $\eprob_{ij}$, and add a negative sign to $e$ with probability $\sprob_{ij}$ (otherwise, the edge has a positive sign). Here, $0\leq \eprob_{ij},\sprob_{ij}\leq 1$ are given probabilities. We refer to graphs sampled in this manner as \textit{inhomogeneous Erd\H{o}s-R\'{e}nyi signed graphs} (see \cref{subsec:random-graph-models}). We then incorporate the edge and sign information into corresponding adjacency and Laplacian matrices, which we denote $A$ and $\mathcal{L}$, respectively. The first half of the paper tackles the following general question: under what conditions on $\eprob_{ij}$ and $\sprob_{ij}$ can we guarantee that, with high probability, $A$ and $\mathcal{L}$ will be ``close'' to their corresponding means? Our main technical contributions consist of two matrix concentration inequalities which, loosely speaking, assert that as long as the probabilities $\eprob_{ij}$ are ``not too sparse,'' and the probabilities $\sprob_{ij}$ are bounded away from $1/2$, then we can guarantee concentration near the mean adjacency and Laplacian matrices in operator norm with high probability. We state below the version for adjacency matrices; the version for Laplacian matrices appears as~\cref{th:laplacian-matrix-spectral}.

    \begin{theorem}[Informal statement of \cref{th:adjacency-matrix}]\label{th:intro-adjacency-matrix}
        Let $A$ be a random signed adjacency matrix on $n$ nodes. Assume $\max_{i, j}|\sprob_{ij} - 1/2| > c_0$ for some $c_0>0$. Assume that
            \begin{align*}
                n\max_{i, j\in[n]} |\eprob_{ij}(1-2\sprob_{ij})| \leq \alpha
            \end{align*}
        for some $\alpha\geq c_1\log{n}$ and $c_1 >0$. Then for any $r>0$ there exists $C = C(r, c_0, c_1) >0$ such that for $n$ sufficiently large and with probability at least $1 - n^{-r}$, it holds
            \begin{align*}
                \op{A - \ol{A}} \leq C\sqrt{\alpha}.
            \end{align*}
    \end{theorem}        

\begin{figure}[t!]
        \begin{center}
            \includegraphics[width=\textwidth]{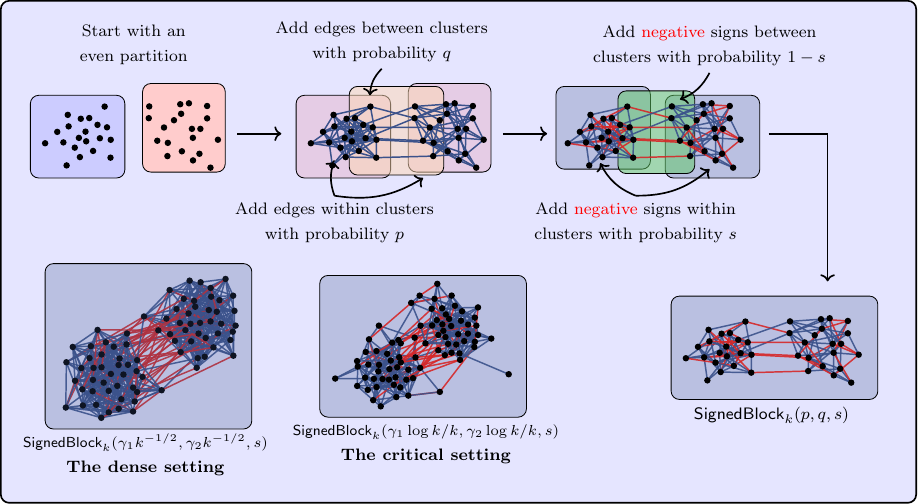}
            \caption{A diagram describing the SSBM in the two-community setting (see \cref{defn:signed-bisection}). }\label{fig:ssbm-diagram}
        \end{center}
    \end{figure}

    Note that this result can be considered a generalization of the strong matrix concentration inequality of Lei and Rinaldo in~\cite{lei2015consistency}. As a result of our technical contributions, in the second half of the paper, we are able to facilitate a detailed study of the spectral properties of a two-way signed stochastic block model (SSBM). We describe this model diagrammatically in \cref{fig:ssbm-diagram}; see \cref{defn:general-ssbm} and \cref{defn:signed-bisection} for a formal definition. Stochastic block models (see, e.g.,~\cite{holland1983stochastic, nowicki2001estimation}) are widely used in statistics and data science to provide principled, generative models for network data constructed from an underlying partition of nodes into communities or blocks. The particular case of a stochastic block model defined with respect to an even bipartition, also called a planted community model, has emerged as a testing ground for community detection and node clustering algorithms. In this paper we consider a signed variant of the traditional stochastic block model by incorporating the presence of negative edges within and between communities with various probabilities. Such models have been studied for their applications to graph clustering and signature synchronization problems (see, e.g., ~\cite{jiang2015stochastic, he2022sssnet, he2022gnns}).

    Our treatment focuses on the two-community setting where edges and signs are generated as follows. Starting with an even partition of $[2k]$, we add edges with endpoints in the same community with probability $p$, and edges with endpoints between the communities with probability $q$. Similarly, we add negative signs to edges with endpoints in the same community with probability $\sprob$, and we add negative signs to edges with endpoints in different communities with probability $1-\sprob$. Here, $p, q\in (0, 1)$ are taken such that $p > q$ and depend on the community size $k$, and $\sprob\in(0, 1/2)$ is taken to be independent of $k$. If $A$ is a signed adjacency matrix obtained in this manner, we write $A\sim\signedblock_k(p, q, \sprob)$. We consider two cases of growth rates for $p, q$ in the form of (1), where $p, q = \Theta(k^{-1/2})$, which we term the ``dense setting'' and (2), where $p, q = \Theta(\log{k}/k)$, which we term the ``sparse setting.'' Notably, in both of these settings, the spectral gap $\lambda_1(\mathcal{L})$ concentrates near $2\sprob$ with high probability. We state this informally below.

    \begin{figure}[t!]
        \begin{center}
            \begin{subfigure}{\textwidth}
                \begin{center}
                    \includegraphics{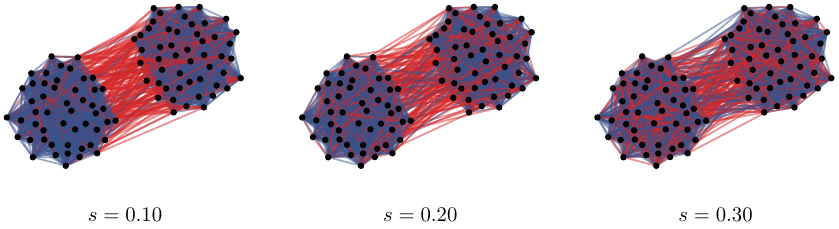}
                \end{center}
            \end{subfigure}\subcaption{Three signed graphs sampled from $\signedblock_{k=50}(10\log{k}/k, \log{k}/k, \sprob)$ for various choices of $\sprob$. Red edges (resp. blue edges) are negatively signed (resp. positively signed).}
            \vspace*{0cm}
            \begin{subfigure}{\textwidth}
                \begin{center}
                    \input{ssbm_picture_hist_1}
                \end{center}
            \end{subfigure}\subcaption{Eigenvalue histograms corresponding to the normalized signed Laplacian matrices of the three graphs shown in (a). We note the correspondence between the least eigenvalue of each graph and $2\sprob$.}
        \end{center}
        \vspace*{-.6cm}
        \caption{An illustration of graphs sampled from the SSBM and their corresponding Laplacian eigenvalue distributions.}\label{fig:samples-and-histograms}
    \end{figure}    

    \begin{theorem}[Informal statement of \cref{cor:eigvals-concentration-lap-dense} and \cref{cor:eigvals-concentration-lap}]\label{cor:intro-eigvals-concentration-lap}
        Suppose\break $A\sim\signedblock_k(p, q, \sprob)$ is a random signed adjacency matrix with associated normalized signed Laplacian matrix $\mathcal{L}$. Let $\lambda_1=\lambda_1(\mathcal{L})$ denote the least eigenvalue of $\mathcal{L}$. Then there exists constants $C, C'>0$, not depending on $k$, such that for $k$ sufficiently large and with probability at least $1-k^{-\Omega(1)}$, the following eigenvalue estimates hold:   
            \begin{enumerate}[label={\normalfont(\arabic*)}]
                \item In the dense setting, $|\lambda_1-2\sprob| \leq C k^{-1/4}$.
                \item In the sparse setting, $|\lambda_1-2\sprob| \leq C'{(\log{k})}^{-1/2}$.
            \end{enumerate}
    \end{theorem}

    \cref{fig:samples-and-histograms} demonstrates this phenomenon for a few choices of $\sprob$. Moreover, the corresponding eigenvector concentrates near the mean Laplacian eigenvector which is given by $u = \frac{1}{\sqrt{2k}}\begin{bmatrix} \mathbf{1}_k & -\mathbf{1}_k \end{bmatrix}^T$ (see \cref{lem:properties}). We state this as an informal theorem below.

    \begin{theorem}[Informal statement of \cref{cor:eigvals-concentration-lap-dense} and \cref{cor:eigvals-concentration-lap}]\label{thm:intro-eigvals-concentration-lap}
        Suppose\break $A\sim\signedblock_k(p, q, \sprob)$ is a random signed adjacency matrix with associated normalized signed Laplacian matrix $\mathcal{L}$. Let $u_1$ denote an eigenvector of $\mathcal{L}$ corresponding to $\lambda_1(\mathcal{L})$. Then there exists constants $C, C'>0$ such that for $k$ sufficiently large and with probability at least $1-k^{-\Omega(1)}$, $\lambda_1$ is simple and there exists a choice of sign $\tau \in \{\pm1\}$ such that the following holds:
            \begin{enumerate}[label={\normalfont(\arabic*)}]
                \item In the dense setting, $\two{\tau u_1 - u} \leq C k^{-1/4}$.
                \item In the sparse setting, $\two{\tau u_1 - u}  \leq C'{(\log{k})}^{-1/2}$.
            \end{enumerate}
    \end{theorem}

    Functionally, \cref{thm:intro-eigvals-concentration-lap} shows that eigenvector estimators of the balanced community detection and $\pm 1$ node synchronization problems are statistically consistent in a weak sense (see~\cref{rmk:consistency}). We remark that much of the theoretical techniques in this article appear robust to generalizations to the case of graphs with edges by elements of $U(1)$, also known as magnetic graphs (see, e.g., ~\cite{lange2015frustration}). Our findings in the $\pm 1$ case suggest future work on statistical consistency of spectral approaches to the $U(1)$ synchronization problem (see \cref{rmk:magnetic}).

    \subsection{Related work}\label{subsec:related-work}


    Earlier references on random signed graphs and balance theory include~\cite{frank1979balance,el2012balance}, and earlier work concerning signed variants of the stochastic block model include~\cite{jiang2015stochastic,yang2017stochastic}. Statistical analyses of SSBM appear in~\cite{wang2022exact,cucuringu2019sponge}. We note that~\cite{cucuringu2021regularized} includes matrix concentration results of a similar nature to those presented in this article, with an important distinction being the authors' usage of a regularized Laplacian in their community detection method and subsequent theoretical analysis, and our usage of stronger adjacency matrix concentration techniques. General graph adjacency and Laplacian matrix concentration inequalities appear in such works as~\cite{oliveira2009concentration,chung2011spectra,lei2015consistency,deng2021strong,eldridge2018unperturbed}. This article utilizes and adapts arguments which appear in~\cite{lei2015consistency,deng2021strong} and extends them to the signed setting (and others, see \cref{rmk:magnetic}).

    Theoretical analyses of signed graphs in the context of balanced community detection, the synchronization problem, and Cheeger inequalities include, for example,~\cite{singer2011angular,wang2013exact,bandeira2013cheeger,lange2015frustration,atay2020cheeger}. Signed graphs have a great degree of interdisciplinary interest, including biology~\cite{granell2011mesoscopic,mason2009signed}, computer science and machine learning~\cite{facchetti2012exploring, tang2016survey,derr2018signed,huang2019signed}, and the social sciences~\cite{szell2010multirelational,traag2009community,yang2007community,leskovec2010signed}.

    \subsection{Outline of this article}

    In \cref{sec:background} we establish our notation and cover relevant background from graph theory and related areas. In \cref{sec:concentration-results} we cover matrix concentration inequalities for signed adjacency (\cref{subsec:concentration-adj}) and Laplacian matrices (\cref{subsec:concentration-lapl}). In \cref{sec:ssbm} we focus on the SSBM and develop results concerning the spectral properties of signed adjacency and Laplacian matrices in the dense setting (\cref{subsec:dense-setting}) and sparse setting (\cref{subsec:sharp-setting}). In \cref{sec:experimental-data}, we present a selection of experimental results showing our theoretical contributions in action. Finally, in \cref{sec:missing-proofs}, we include proofs of various results which appear throughout the preceding sections of the paper.

    \section{Background}\label{sec:background}

    In this section we establish our notational conventions and cover relevant background topics. We divide the section into two pieces, the first concerning signed graph theory and related concepts (\cref{subsec:signed-graphs}), and the second concerning random graph models (\cref{subsec:random-graph-models}).

    \subsection{Signed graphs, community detection, and synchronization}\label{subsec:signed-graphs}

    We consider graphs of the form $G= (V, E)$, where $V = [n]$ and $E\subseteq {[n]\choose 2}$; that is, $G$ is finite, undirected, and has no loops or multiple edges. We write $i\sim j$ whenever $\{i, j\}\in E$ and denote by $d_i$ the \textit{degree} of node $i$, which is the number of edges incident to $i$.

    A \textit{signature} is a map $\signature:E\rightarrow\{\pm 1\}$ which assigns to each edge a sign which is either positive or negative. We say that $(G, \signature)$ is a \textit{signed graph}. The \textit{signed adjacency matrix} of a graph $G$ is the matrix $A = (a_{ij})\in\mathbb{R}^{n\times n}$ with entries
        \begin{align*}
            a_{ij} &= \begin{cases}
                \signature_{ij} &\text{ if } i\sim j\\
                0&\text{ otherwise}
            \end{cases},\hspace{.5cm}i, j\in V.
        \end{align*}
    Let $D = \mathrm{diag}(d_1, \dotsc, d_n)\in\mathbb{R}^{n\times n}$ be the diagonal matrix of node degrees. The \textit{signed Laplacian matrix} is defined by $L = D-A$. The \textit{normalized signed Laplacian matrix} is given by $\mathcal{L} = D^{-1/2}LD^{-1/2}$ where we abuse notation and set $D^{-1/2}_{ii} = 0$ whenever $d_{i} = 0$. The entries of $\mathcal{L}= (l_{ij})_{i,j=1}^{n} $ can be expressed in the form
        \begin{align*}
            l_{ij} &= \begin{cases}
                1&\text{ if }i=j\\
                -\frac{\signature_{ij}}{\sqrt{d_i d_j}} &\text{ if } i\sim j\\
                0&\text{ otherwise}
            \end{cases},\hspace{.5cm}i, j\in [n].
        \end{align*}

    In this article we suppress the dependence of $A, L, \mathcal{L}$ on $\signature$ since we do not encounter scenarios in which multiple signatures may play a role. We use $|A|$ to refer to the (unsigned) adjacency of the underlying graph, and in general $|\cdot|$ to refer to the \textit{entrywise} absolute value of a given matrix. Note that $L$ and $\mathcal{L}$ are both positive semidefinite and possess a complete set of nonnegative eigenvalues with corresponding eigenvectors. We note that unlike in the case of traditional graph Laplacians, it need not hold that $\lambda_1(\mathcal{L}) =0$.
    
    When $X\in\mathbb{R}^{n\times n}$ is any symmetric matrix, we denote its eigenvalues, ordered in ascending fashion, as follows $\lambda_1(X)\leq \lambda_2(X)\leq\dotsc\leq \lambda_n(X)$. When $X$ is any matrix, we denote by $\op{X}$ the $\ell_2$-operator or spectral norm of $X$ and by $\fro{X}$ the Frobenius norm of $X$. We denote by $\two{x}$ the $\ell_2$ norm of a vector $x$.

    If $f:V\rightarrow\mathbb{R}$ is any function, we can define its \textit{$\ell_2$-frustration} by the following expression
        \begin{align*}
            \frustration_2(f) &= \frac{\sum_{\{i, j\}\in E} |f(i)-\signature_{ij}f(j)|^2 }{\sum_{i\in V} |f(i)|^2 d_i}
        \end{align*}
    The frustration of a function can be understood as a measure of how well $f(i)$ satisfies the signs of each edge. A signature $\signature$ is said to be \textit{balanced} or \textit{consistent} whenever there exists a function $f:V\rightarrow\{\pm 1\}$ such that $\signature_{ij} = f(i)f(j)^{-1}$ for each $\{i, j\}\in E$, i.e., such that $\frustration_2(f)=0$. We define the \textit{$\ell_2$-frustration index of $\signature$} as the minimum
        \begin{align}\label{eq:frustration-index}
            \frustration_2(\signature) &= \min_{f:V\rightarrow\{\pm 1\}} \frustration_2(f).
        \end{align}
    The frustration index has been studied in more general settings which include $\mathsf{U}(1)$ and $\mathsf{O}(d)$ labellings of the edges, which are often called magnetic and connection graphs, respectively (see, e.g.,~\cite{lange2015frustration,singer2012vector}). Note that $\frustration_2(\signature)$ is related to $\lambda_1(\mathcal{L})$ by a semidefinite relaxation of~\cref{eq:frustration-index} via the variational characterization of $\lambda_1(\mathcal{L})$, namely,
        \begin{align*}
            \lambda_1(\mathcal{L}) &= \inf_{\substack{f:V\rightarrow\mathbb{R} \\ f\neq 0}} \frac{f^T \mathcal{L} f}{ f^T f} = \inf_{\substack{f:V\rightarrow\mathbb{R} \\ f\neq 0}} \frac{\sum_{\{i, j\}\in E} |f(i)-\signature_{ij}f(j)|^2 }{\sum_{i\in V} |f(i)|^2 d_i}.
        \end{align*}
    In~\cite{bandeira2013cheeger}, the authors show that $\lambda_1(\mathcal{L})$ and $\frustration_2(\signature)$ are related by a Cheeger-type inequality of the form $\frustration_2(\signature)\leq \sqrt{8 \lambda_1(\mathcal{L})}$. Spectral lower bounds can also obtained in the setting where $\frustration_2(\signature)$ is slightly relaxed to cover $f:V\rightarrow\{\pm 1, 0\}$. This bound is obtained through an algorithmic approach to solving the $\{\pm 1, 0\}$ relaxation by obtaining the first eigenvector $u_1$ of $\mathcal{L}$ and using it to construct a function $f^\ast$ by modifying $\mathrm{sgn}(u_1)$ whose frustration is provably proximal to the frustration of $\signature$. Here, if $x\in\mathbb{R}^n$ is any vector, $\mathrm{sgn}(x)\in\mathbb{R}^n$ is the vector given by $\mathrm{sgn}(x)_i = 1$ if $x_i \geq 0$ and $\mathrm{sgn}(x)_i = -1$ otherwise. We mention in \cref{rmk:consistency} that in the two-community SSBM setting, $\mathrm{sgn}(u_1)$ defines a statistically consistent estimator of $\frustration_2(\signature)$ in a weak sense. 
    
    Signature frustration and the eigenvalues of $\mathcal{L}$ are also related to the edge expansion properties of $G$ and can thus also be used to study community detection problems on signed networks. As an example of this approach, define for $V_1\subseteq V$ the {\textit{$\ell_1$-frustration}} according to
        \begin{align*}
            \frustration_1(V_1) = \min_{f:V_1\rightarrow\{\pm1\}} \sum_{\{i, j\}\in E(V_1)} |f(i) - \signature_{ij} f(j)|.
        \end{align*}
    We note that, as was pointed out in~\cite{lange2015frustration}, $\frustration_1(V)$ is equal to the minimum number of edges which need be removed from $E$ in order to ensure the resulting graph is balanced. One may then define the {\textit{$\ell_1$-frustration index of $\signature$}} according to
        \begin{align*}
            \frustration_1(\signature) &=\min_{{\varnothing \neq V_1 \subseteq V}} \frac{\frustration_1(V_1) + |E(V_1, V_1^c)|}{\sum_{i\in V_1}d_i}
        \end{align*}
    The program $\frustration_1(\signature)$ can be thought of as searching for a subset of $V$ which has ideal edge expansion and balance properties; in the community detection literature such problems are often described as finding balanced communities~\cite{anchuri2012communities,ordozgoiti2020finding}. Cheeger-type inequalities related to $\frustration_1(\signature)$ are also studied extensively in~\cite{lange2015frustration} (with a notable distinction from our case being the authors' usage of a non-normalized Laplacian matrix). The authors obtain, for example,
        \begin{align*}
            \frac{1}{2}\lambda_1(L) \leq \frustration_1(\signature) \leq \sqrt{8\Delta \lambda_1(L)},
        \end{align*}
    where $\Delta$ is the maximum degree of the graph $G$. It is within the context of these interpretations of $\lambda_1$ that the concentration $\lambda_1(\mathcal{L})\approx 2\sprob$ (as described in \cref{thm:intro-eigvals-concentration-lap}) suggests an understanding of $\lambda_1$ as a probabilistic quantity related to misappropriated edge signatures.

    \subsection{Random graphs and the SSBM}\label{subsec:random-graph-models}

    In this subsection we cover background on random graph models which will be of use to us in the sections to follow. We begin by letting $\eprmatrix = (\eprob_{ij})_{i, j=1}^n\in\mathbb{R}^{n\times n}$ be a symmetric matrix of probabilities, i.e., such that $\eprob_{ij}\in[0, 1]$ for all $i, j$ and $\eprob_{ij} = \eprob_{ji}$. We assume $\eprob_{ii} = 0$ for all $i$. We can construct a random graph on $[n]$ by adding an edge $\{i, j\}$ with probability $\eprob_{ij}$, for all such $\{i, j\}\in{[n]\choose 2}$, with the requirement that all pairs are sampled in a mutually independent fashion. A graph which has been constructed according to this model is said to be sampled from the distribution $\mathsf{G}_n(\eprmatrix)$. Note that if $\eprob_{ij} = p$ for some $p\in[0, 1]$ independent of $i, j$ then $\mathsf{G}_n(\eprmatrix)$ is simply the well-known Erd\H{o}s-R\'{e}nyi random graph model. The model $\mathsf{G}_n(\eprmatrix)$ is sometimes called the \textit{inhomogeneous Erd\H{o}s-R\'{e}nyi} model~\cite{oliveira2009concentration}. We recall the following concentration inequality of Lei and Rinaldo~\cite{lei2015consistency} for random adjacency matrices drawn from $\mathsf{G}_n(\eprmatrix)$, off of which one of our technical contributions is based, and which will be used directly later on in the proof of \cref{th:adjacency-matrix} (see \cref{subsec:proofs-concentration}). 

    \begin{theorem}[Lei and Rinaldo,~\cite{lei2015consistency,supplei2015consistency}]\label{thm:unsigned-adjacency}
        Let $A\sim \mathsf{G}_n(\eprmatrix)$ be a random adjacency matrix on $n$ nodes. Assume that $\max_{i, j} \eprob_{ij} \leq \alpha/n$ for some $\alpha\geq c_0\log{n}$ and $c_0 >0$. Then for any $r>0$ there exists $C = C(r, c_0) >0$ such that for $n$ sufficiently large and with probability at least $1 - n^{-r}$, it holds
            \begin{align*}
                \op{A - \ol{A}} \leq C\sqrt{\alpha}.
            \end{align*}
    \end{theorem}

    In this article we consider an extension of the inhomogeneous Erd\H{o}s-R\'{e}nyi model to signed graphs. Let $\sprmatrix = (\sprob_{ij})\in\mathbb{R}^{n\times n}$ be another symmetric matrix of probabilities, again taking $\sprob_{ii} = 0$ for all $i$. Now having fixed a pair $(\eprmatrix,\sprmatrix)$ construct a random signed graph as follows. For each pair $\{i, j\}\in{[n]\choose 2}$, add an edge with probability $\eprob_{ij}$ and add a negative sign with probability $\sprob_{ij}$. All pairs and all signs are sampled mutually independently. A signed graph constructed in this manner is to have been sampled from the distribution $\mathsf{G}_n(\eprmatrix,\sprmatrix)$, and we will refer to this as the \textit{signed inhomogeneous Erd\H{o}s-R\'{e}nyi} model. If this is the case, we write $A\sim \mathsf{G}_n(\eprmatrix,\sprmatrix)$. 

    If $A\sim \mathsf{G}_n(\eprmatrix,\sprmatrix)$, then we may write $a_{ij} = P_{ij}(1-2S_{ij})$ where $P_{ij}$ is a Bernoulli random variable with probability $\eprob_{ij}$ and $S_{ij}$ is a Bernoulli random variable with probability $\sprob_{ij}$. We write $\overline{A} = (\overline{a}_{ij})\in\mathbb{R}^{n\times n}$ to denote the mean signed adjacency matrix for the model $\mathsf{G}_n(\eprmatrix,\sprmatrix)$ (we suppress the dependence of $\ol{A}$ on $\eprmatrix,\sprmatrix$ for brevity; the probability matrices will be clear in context). That is, 
        \begin{align}\label{eq:mean-signed-adjacency}
            \overline{a}_{ij} = \eprob_{ij}(1-2\sprob_{ij}).
        \end{align}
    We denote by $\overline{d}_i$ the expected degree of node $i$, and the matrix $\overline{D}$ will denote the diagonal matrix of expected degrees. Concretely, $\overline{d}_i = \sum_{j=1}^n \eprob_{ij}$. The matrix $\overline{L}$ is the mean signed Laplacian matrix, and can be expressed $\overline{L} = \overline{D} - \overline{A}$. The matrix $\ol{\mathcal{L}} = \ol{D}^{-1/2}\ol{L}\ol{D}^{-1/2}$ is the degree-normalized mean Laplacian matrix.
    
    We now introduce our main model of interest, the SSBM. We start with the general construction. We denote by $J_{s\times t}\in\mathbb{R}^{s\times t}$ the matrix containing all ones. When $s=t$, $I_{s\times s}\in\mathbb{R}^{s\times s}$ denotes the identity matrix, and $\widetilde{J}_{s\times s} = {J}_{s\times s} - {I}_{s\times s}$ is the matrix $J_{s\times s}$ with zero diagonal. 

    \begin{definition}[Signed stochastic block model]\label{defn:general-ssbm}
        Let $n\geq 2$ be fixed, and let\break $Q_1, Q_2, \dotsc, Q_{m}$ be a pairwise disjoint and nonempty $m$-partition of the integers $[n]$, where $Q_i\subseteq [n]$ for each $i$ and $m\geq 2$. Let the matrix $P=(\eprob_{st})_{s,t=1}^{m}\in\mathbb{R}^{m\times m}$ consist of the probabilities that an edge is added between each pair of clusters, i.e.,
            \begin{align*}
                \eprob_{st} = \pr{\text{an edge is added between }i\in Q_s\text{ and }j\in Q_t}\in [0, 1],\hspace{.25cm}1\leq s, t\leq m.
            \end{align*}
        Similarly, let the matrix $S=(\sprob_{st})_{s,t=1}^{m}\in\mathbb{R}^{m\times m}$ consist of the probabilities that a negative sign is added between each pair of clusters, i.e.,
            \begin{align*}
                \sprob_{st} = \pr{\text{an negative sign is added between }i\in Q_s\text{ and }j\in Q_t}\in [0, 1],\hspace{.25cm}1\leq s, t\leq m.
            \end{align*}
        Now extend these in the following way to $\eprmatrix\in\mathbb{R}^{n\times n}$, writing
            \begin{align*}
                \eprmatrix &= \begin{pmatrix}
                    \eprob_{11}\widetilde{J}_{|Q_1|\times |Q_1|} & \eprob_{12}J_{|Q_1|\times |Q_2|} & \dotsc & \eprob_{1 m}J_{|Q_1|\times |Q_m|}\\
                    \eprob_{21}J_{|Q_2|\times |Q_1|} & \eprob_{22}\widetilde{J}_{|Q_1|\times |Q_1|} & \dotsc & \eprob_{2 m}J_{|Q_2|\times |Q_m|}\\
                    \vdots & \vdots & \ddots & \vdots\\
                    \eprob_{m1}J_{|Q_m|\times |Q_1|} &  \eprob_{m2}J_{|Q_m|\times |Q_2|} & \dotsc & \eprob_{mm}\widetilde{J}_{|Q_m|\times |Q_m|}
                \end{pmatrix}.
            \end{align*}
        Define $\sprmatrix\in\mathbb{R}^{n\times n}$ in the same manner. Then the corresponding inhomogeneous signed Erd\H{o}s-R\'{e}nyi random graph model $\mathsf{G}_n(\eprmatrix,\sprmatrix)$ is called {\normalfont{the signed stochastic block model (SSBM)}}.
    \end{definition}

    We consider, as follows, a specialization of this model to the case consisting of two communities and with signs determined by a single parameter $\sprob$ (see \cref{fig:ssbm-diagram}).

    \begin{definition}[Signed stochastic bisection model]\label{defn:signed-bisection}
        Let $k\geq 2$, and begin with the even bipartition $[2k] = [k]\sqcup ([2k]\setminus [k])$. Let $p, q, \sprob \in(0, 1)$ be fixed parameters. Define the matrices $P_{p, q}, S_{\sprob}\in\mathbb{R}^{2\times2}$ as follows:
            \begin{align*}
                P_{p, q} =\begin{pmatrix}
                    p & q\\
                    q & p
                \end{pmatrix}, \hspace{.25cm} S_{\sprob} =\begin{pmatrix}
                    \sprob & 1-\sprob\\
                    1-\sprob & \sprob
                \end{pmatrix}.
            \end{align*}
        Let $\eprmatrix_{p, q}, \sprmatrix_{\sprob}$ denote the matrices indexed by $[2k]$ obtained by extending $P_{p, q}, S_{\sprob}$ as in \cref{defn:general-ssbm}. We define the {\normalfont{signed stochastic bisection model}} to be given by $\mathsf{G}_{2k}(\eprmatrix_{p, q}, \sprmatrix_{\sprob})$, and which we alternatively denote $\signedblock_k(p, q, \sprob)$.
    \end{definition}

    To conclude this section we provide a lemma which establishes some of the basic properties of $\signedblock_k(p, q, \sprob)$.

    \begin{lemma}\label{lem:properties}
        Let $k\geq 2$ be fixed. Let $p, q, \sprob \in(0, 1)$ be fixed parameters and let \break $A\sim \signedblock_k(p, q, \sprob)$ be a random signed adjacency matrix with corresponding signed Laplacian matrix $\mathcal{L}$. Then we have the following facts:
            \begin{enumerate}[label=\textit{(\roman*)}]
                \item For each $i\in [2k]$, the expected degree is given by:
                    \begin{align*}
                        \ol{d}_i &= p(k-1) + k q =: \ol{d}
                    \end{align*}
                \item The mean adjacency matrix $\ol{A}$ is given by:
                    \begin{align*}
                        \ol{A} = \begin{pmatrix}
                            p(1-2\sprob)\widetilde{J}_{k\times k} & q(2\sprob-1)J_{k\times k}\\
                            q(2\sprob-1)J_{k\times k} & p(1-2\sprob)\widetilde{J}_{k\times k}
                        \end{pmatrix}.
                    \end{align*}
                The eigenvalues of $\ol{A}$ are as follows:
                    \begin{align*}
                        (1-2\sprob)(p(k-1) + qk)&\text{ with multiplicity }1,\\ 
                        (1-2\sprob)(p(k-1) - qk)&\text{ with multiplicity }1,\text{ and }\\ 
                        -p(1-2\sprob),&\text{ with multiplicity }n-2.
                    \end{align*}
                \item The degree-normalized mean laplacian matrix $\ol{\mathcal{L}}$ is given by:
                    \begin{align*}
                        \ol{\mathcal{L}} = \frac{1}{\ol{d}}\begin{pmatrix}
                            \ol{d}I_{k\times k} - p(1-2\sprob)\widetilde{J}_{k\times k} & -q(2\sprob-1)J_{k\times k}\\
                            -q(2\sprob-1)J_{k\times k} & \ol{d}I_{k\times k} - p(1-2\sprob)\widetilde{J}_{k\times k}
                        \end{pmatrix}
                    \end{align*}
                The eigenvalues of $\ol{\mathcal{L}}$ are as follows:
                    \begin{align*}
                        2\sprob &\text{ with multiplicity }1,\\ 
                        1 - (1-2\sprob)\frac{p(k-1) - qk}{p(k-1) + qk}&\text{ with multiplicity }1,\text{ and }\\ 
                        1+(1-2\sprob)\frac{p}{p(k-1) + qk},&\text{ with multiplicity }n-2.
                    \end{align*}
            \end{enumerate}
    \end{lemma}

    A proof of this result is given in \cref{subsec:proofs-background}.

    \section{Concentration for signed adjacency and Laplacian matrices}\label{sec:concentration-results}
    In this section we state and prove the main technical contributions of this paper, which consist of matrix concentration results, respectively, for random signed adjacency and Laplacian matrices.

    \subsection{Signed adjacency matrices}\label{subsec:concentration-adj}
    Our main result is stated as follows. We note that this result is based on and extends the concentration inequality of~\cite{lei2015consistency}, upon whose proof ours is based\footnote{Going a step further, Lei and Rinaldo note that their proof is itself based on the earlier work of Feige and Ofek in~\cite{feige2005spectral}, who study the eigenvalues of the traditional Erd\H{o}s-R\'{e}nyi random graph model.}. We note that the authors' proof appears in a supplement to the original paper, which is published in~\cite{supplei2015consistency}. 

    \begin{theorem}\label{th:adjacency-matrix}
        Let $A\sim \mathsf{G}_n(\eprmatrix,\sprmatrix)$ be a random signed adjacency matrix. 
        Assume \break $\min_{i, j}|\sprob_{ij} - 1/2| > c_0$ for some $c_0>0$. Assume that $n \max_{i, j} |\ol{a}_{ij}| \leq \alpha$ for some $\alpha\geq c_1\log{n}$ and $c_1 >0$. Then for any $r>0$ there exists $C = C(r, c_0, c_1) >0$ such that for $n$ sufficiently large and with probability at least $1 - n^{-r}$, it holds
            \begin{align*}
                \op{A - \ol{A}} \leq C\sqrt{\alpha}.
            \end{align*}
    \end{theorem}

    To bound the distance $\op{A - \ol{A}}$, we focus on controlling the term $|x^T (A-\ol{A})y|$ for $x, y\in\mathbb{R}^n$ with $\two{x} = \two{y}=1$. The first step is to discretize this region as follows. For $t\geq 0$ write $B_{t} = \{x\in\mathbb{R}^n : \two{x} \leq t\}$ and $B_1 =: B$. Then, for $\delta\in(0, 1)$, define the following set $\mathfrak{B}$, which serves as a grid approximation of the ball $B$ with precision $\delta/\sqrt{n}$:
        \begin{align*}
            \mathfrak{B} = \left\{x = (x_1, x_2, \dotsc, x_n) \in S : \sqrt{n} x_i / \delta\in \mathbb{Z}\text{ for each }i\right\}.
        \end{align*}
    We make use of the following lemma, which appears in~\cite{feige2005spectral} and~\cite{lei2015consistency}. We omit a proof.

    \begin{lemma}\label{lemma:discretization}
        $S_{1-\delta}\subseteq\mathrm{convex hull}(\mathfrak{B})$. Moreover, for each $X\in\mathbb{R}^{n\times n}$, it holds
        \begin{align*}
            \op{X}\leq (1-\delta)^{-2}\sup_{x, y\in \mathfrak{B}}|x^T Xy|.
        \end{align*}
    \end{lemma}

    It can be shown via a volume argument that the cardinality $\card{\mathfrak{B}}$ satisfies $\card{\mathfrak{B}} \leq e^{n\log(9/\delta)}$ (see, e.g., \cite[Claim 2.9]{feige2005spectral}). Note that, for $x, y\in\mathbb{R}^{n}$ fixed, it holds
        \begin{align*}
            x^T(A - \ol{A})y &= \sum_{i, j = 1}^n x_i (a_{ij} - \ol{a}_{ij}) y_j = \sum_{i, j = 1}^n x_i w_{ij} y_j.
        \end{align*}
    where we let $W = (w_{ij})_{i, j=1}^{n}\in\mathbb{R}^{n\times n}$ denote the matrix $W=A-\ol{A}$.  We split the indices $(i, j)\in [n]\times [n]$ into two complementary sets, as follows:
        \begin{align*}
            \mathfrak{L}(x, y) &= \left\{(i, j) : |x_i y_j| \leq \sqrt{\alpha} /n \right\}, \hspace{.25cm}\text{the ``light pairs,'' and}\\
            \ol{\mathfrak{L}}(x, y) &= \left\{(i, j) : |x_i y_j| > \sqrt{\alpha} /n \right\}, \hspace{.25cm}\text{the ``heavy pairs.''}
        \end{align*}
    Therefore, since we have
        \begin{align*}
            \sup_{x, y\in \mathfrak{B}}|x^T W y| \leq \sup_{x, y\in \mathfrak{B}}\left|\sum_{(i, j)\in \mathfrak{L}(x, y)}x_i w_{ij} y_j \right| + \sup_{x, y\in \mathfrak{B}}\left|\sum_{(i, j)\in \ol{\mathfrak{L}}(x, y)}x_i w_{ij} y_j \right|
        \end{align*}
    it suffices to separately bound $\mathfrak{L}(x, y)$ and $\ol{\mathfrak{L}}(x, y)$. First we handle the light pairs, as follows.

    \begin{lemma}\label{lemma:light-pairs}
        Instate the hypotheses of \cref{th:adjacency-matrix}. For each $t>0$, we have
            \begin{align*}
                \pr{\sup_{x, y\in \mathfrak{B}} \left|\sum_{(i, j)\in\mathfrak{L}(x, y)} x_i w_{ij} y_j\right| \geq t\sqrt{\alpha}} \leq 2\exp\left\{ -n\left(\frac{t^2}{\frac{2}{c_0} + \frac{8t}{3}} - 2\log(9/\delta)\right) \right\}.
            \end{align*}
    \end{lemma}
    
    We include a proof in \cref{subsec:proofs-concentration}. Having bounded the light pairs, we turn to the complementary set of indices and provide a bound as follows.

    \begin{lemma}\label{lemma:heavy-pairs}
        Instate the hypotheses of \cref{th:adjacency-matrix}. Then there exists $C>0$ depending only on $r, c_0$ such that
            \begin{align*}
                \pr{\sup_{x, y\in \mathfrak{B}}\left|\sum_{(i, j)\in\ol{\mathfrak{L}}(x, y)} x_i y_j w_{ij}\right| \geq C\sqrt{\alpha}}\leq n^{-r}.
            \end{align*}
    \end{lemma}

    The proof is given in \cref{subsec:proofs-concentration} and relies on a lemma (namely,~\cite[Lemma 4.3]{supplei2015consistency}) which appears in the proof of the unsigned concentration inequality \cref{thm:unsigned-adjacency}. Finally, we put the ingredients together to prove the main claim.

    \begin{proof}[Proof of \cref{th:adjacency-matrix}]
        Letting $\delta\in(0, 1)$ be fixed, by \cref{lemma:discretization}, we have
            \begin{align}\label{eq:initial-estimate}
                \op{A-\ol{A}}&\leq(1-\delta)^{-2} \sup_{x, y\in\mathfrak{B}} |x^T (A-\ol{A})y|\\
                &\leq \sup_{x, y\in\mathfrak{B}} \left|\sum_{i,j\in\mathfrak{L}(x,y)} x_i y_j w_{ij}\right| +  \sup_{x, y\in\mathfrak{B}} \left|\sum_{i,j\in\ol{\mathfrak{L}}(x,y)} x_i y_j w_{ij}\right| 
            \end{align}
        By \cref{lemma:heavy-pairs}, there exists $C = C(r, c_0)>0$ such that with probability at least $1-n^{-r}$, 
            \begin{align*}
                \sup_{x, y\in\mathfrak{B}} \left|\sum_{i,j\in\ol{\mathfrak{L}}(x,y)} x_i y_j w_{ij}\right| &\leq C\sqrt{\alpha}
            \end{align*}
        By \cref{lemma:light-pairs} with $s=C$ and taking, e.g., $\delta=9n^{-1/2}$, we have that with probability at least
            \begin{align*}
                1-2\exp\left\{ -C'n + \log{n} \right\}\geq 1- 2\exp\left\{-\frac{C'}{2}n\right\},
            \end{align*}
        where $C'=2\frac{C^2}{\frac{2}{c_0} + \frac{8C}{3}}$ for brevity, it holds
            \begin{align*}
                \sup_{x, y\in\mathfrak{B}} \left|\sum_{i,j\in\mathfrak{L}(x,y)} x_i y_j w_{ij}\right| \leq C\sqrt{\alpha}.
            \end{align*}
        For $n$ sufficiently large, it holds $1- 2\exp\left\{-\frac{C'}{2}n\right\} \geq 1 - n^{-r}$ and therefore with probability at least $1-2n^{-r}$, it follows that both terms in \cref{eq:initial-estimate} are at most $C\sqrt{\alpha}$ so that
            \begin{align*}
                \op{A-\ol{A}}\leq 2C\sqrt{\alpha},
            \end{align*}
        as claimed.
    \end{proof}

    \begin{remark}\label{rmk:magnetic}\normalfont
        We remark that the proofs of \cref{th:adjacency-matrix} and its associated lemmata extend to the case where the random signature $S_{ij}$ is allowed to take values in the circle $U(1)$ as an extension of $\{\pm 1\}$, provided one impose some control on $|\ep{S_{ij}}|$ from below. The proof would be identical other than some small modifications.
    \end{remark}

    \subsection{Signed Laplacian matrices}\label{subsec:concentration-lapl}
    In this subsection we extend \cref{th:adjacency-matrix} to the case of signed Laplacian matrices.

    \begin{theorem}\label{th:laplacian-matrix-spectral}
        Let $A\sim \mathsf{G}_n(\eprmatrix,\sprmatrix)$ be a random signed adjacency matrix, and let $\mathcal{L}$ be its associated normalized signed Laplacian matrix. Assume $\min_{i, j}|\sprob_{ij} - 1/2| > c_0$ for some $c_0>0$, and that $n \max_{i, j} |\ol{a}_{ij}| {\leq} \alpha$ for some $\alpha\geq c_1\log{n}$ and $c_1 > 0$. Then for any $r>0$ there exists $C = C(c_0, c_1, r) >0$ such that for $n$ sufficiently large and with probability at least $1 - n^{-r}$, it holds
            \begin{align}
                \op{\mathcal{L} - \ol{\mathcal{L}}}\leq\frac{C \alpha^{5/2}}{\min\left\{d_{\min}, \ol{d_{\min}}\right\}^3}
            \end{align}
        where $d_{\min} = \min(|A|\mathbf{1}_n)$ (resp. $\ol{d_{\min}} = \min(\eprmatrix\mathbf{1}_n$)) is the minimum degree of $A$ (resp. the minimum expected degree of $A$).
    \end{theorem}

    Note that this bound depends on the random variable $d_{\min}$ and thus, in practice, needs to be de-randomized based on the model in question. Our proof is based on that of Deng, Ling, and Strohmer for unsigned random Laplacian matrices as it appears in \cite[Theorem 4]{deng2021strong}, and is included in \cref{subsec:proofs-concentration}.
    
    \section{Application: Spectral properties of the SSBM}\label{sec:ssbm}

    In this section we turn to a spectral and statistical analysis of the SSBM, with our focus on the two-community model $\signedblock_k(p, q, \sprob)$ as introduced in \cref{defn:signed-bisection}. We focus on two settings for this model, first where $p, q = \Theta(k^{-1/2})$ and which we term the \textit{dense setting} and then where $p, q = \Theta(\log{k}/k)$, which we term the \textit{sparse setting}. The terminology sharp refers to the fact $p=\log{n}/n$ is a sharp threshold for connectivity in the traditional Erd\H{o}s-R\'{e}nyi random graph model (see~\cite{erd6s1960evolution}).

    \subsection{The dense setting}\label{subsec:dense-setting}
    We first investigate the convergence of eigenvalues and eigenvectors of the SSBM in a dense setting, constructed as follows. Letting $k\geq 2$, $\gamma_1>\gamma_2>0$, and $0<\sprob<1/2$ be fixed, we consider the case of $\signedblock_k(\gamma_1 k^{-1/2}, \gamma_2 k^{-1/2}, \sprob)$. Based on a comparison to the traditional Erd\H{o}s-R\'{e}nyi random graph model, whenever, e.g., $\gamma_2>1$ the underlying graph of $\signedblock_k(\gamma_1 k^{-1/2}, \gamma_2 k^{-1/2}, \sprob)$ will be connected with probability at least $1-o(1)$ (see, e.g.,~\cite{van2024random} for discussions of the connectivity properties of inhomogeneous random graph models). We first introduce results which concern signed adjacency matrices in this setting, and then cover signed Laplacian matrices.

    \begin{proposition}\label{prop:dense-convergence-adj}
        Let $k\geq 2$ and fix $\gamma_1 > \gamma_2 > 0$, as well as $0<\sprob<1/2$. Let\break $A\sim\signedblock_k(\gamma_1 k^{-1/2}, \gamma_2 k^{-1/2}, \sprob)$. Then there exists $C>0$, such that for $k$ sufficiently large and with probability at least $1-k^{-2}$, it holds
            \begin{align*}
                \op{A-\ol{A}} \leq Ck^{1/4}.
            \end{align*}
    \end{proposition}

    This proposition follows directly from \cref{th:adjacency-matrix} and we thus omit a proof. By Weyl's inequality and \cref{lem:properties}, we have the following eigenvalue concentration inequalities.

    \begin{corollary}\label{cor:eigvals-adjacency}
        Instate the hypotheses of \cref{prop:dense-convergence-adj} and let\break $A\sim\signedblock_k(\gamma_1 k^{-1/2}, \gamma_2 k^{-1/2}, \sprob)$. Then there exists $C>0$ such that with probability at least $1-k^{-2}$, the following eigenvalue concentration inequalities hold:
            \begin{align*}
                |\lambda_{2k}(A) - (1-2\sprob)(\gamma_1+\gamma_2)\sqrt{k}| &\leq Ck^{1/4} + o(1)\\
                |\lambda_{2k-1}(A) - (1-2\sprob)(\gamma_1-\gamma_2)\sqrt{k}| &\leq Ck^{1/4} + o(1)\\
                |\lambda_\ell(A)| &\leq Ck^{1/4} + o(1),\hspace{.25cm}\ell < 2k-1.
            \end{align*}
    \end{corollary}

    We give a short proof for completeness in \cref{subsec:proofs-ssbm}. We note that for $k$ sufficiently large such that $\gamma_1 > \frac{k}{k-1}\gamma_2$, coupled with the requirement $0<\sprob<1/2$, ensure that the ordering of the eigenvalues of $\ol{A}$ are as claimed in \cref{cor:eigvals-adjacency}. To investigate the concentration of the corresponding Laplacian matrix, we first need to derandomize the bound in \cref{th:laplacian-matrix-spectral}. This is done as follows.

    \begin{lemma}\label{lem:min-degree-dense}
        Let $k\geq 2$ and fix $\gamma_1 > \gamma_2>0$, as well as $0<\sprob<1/2$. Let\break $A\sim\signedblock_k(\gamma_1 k^{-1/2}, \gamma_2 k^{-1/2}, \sprob)$. Then there exists $C(\gamma_1,\gamma_2)$ such that, with probability at least $1-2ke^{-k^{1/6}C(\gamma_1,\gamma_2)}$, it holds
            \begin{align*}
                d_{\min} \geq \frac{\gamma_1+\gamma_2}{2}k^{1/2}.
            \end{align*}
    \end{lemma}

    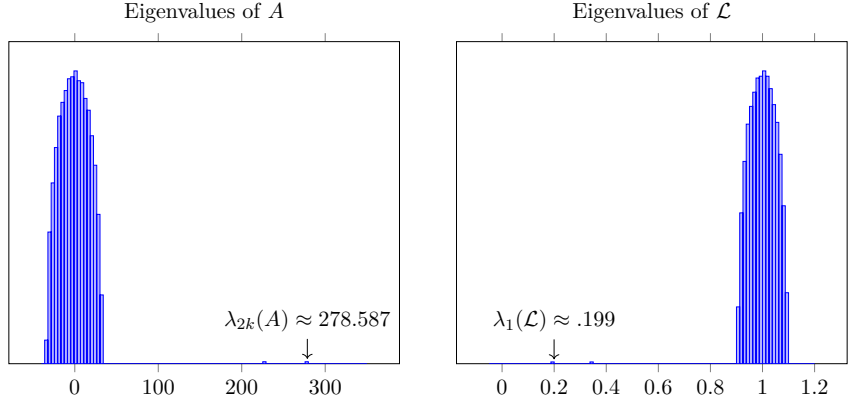
\begin{figure}
        \begin{center}
            \begin{tikzpicture}[scale=.75]
                \begin{axis}[
                    ytick = \empty,
                    name = plot00,
                    ybar,
                    ymin=0,
                    title={Eigenvalues of $A$}
                ]
                \addplot +[
                    hist={
                        bins=100,
                        data min=-40,
                        data max=350
                    },
                    fill=blue!30,
                    draw=blue
                ] table [y index=0] {eigvals_adj_dense_01.csv};
                \draw [<-] (axis cs:278.5873, 2.5)-- +(0pt,10pt) node[above] {$\lambda_{2k}(A)\approx 278.587$};
                \end{axis}
                \begin{axis}[
                    ytick = \empty,
                    name = plot01,at={(plot00.east)},anchor=west,xshift=1cm,
                    ybar,
                    ymin=0,
                    title={Eigenvalues of $\mathcal{L}$}
                ]
                \addplot +[
                    hist={
                        bins=100,
                        data min=-0.05,
                        data max=1.2
                    },
                    fill=blue!30,
                    draw=blue
                ] table [y index=0] {eigvals_lap_dense_01.csv};
                \draw [<-] (axis cs:0.1997, 2.5)-- +(0pt,10pt) node[above] {$\lambda_1(\mathcal{L})\approx .199$};
                \end{axis}
            \end{tikzpicture}
        \end{center}
        \caption{A histogram of the eigenvalues of an observed random signed adjacency matrix $A$ and Laplacian matrix $\mathcal{L}$ obtained from the dense setting. Here, we take $k=10^3$, $\gamma_1=10$, $\gamma_2=1$, $\sprob=0.1$. In this case, $\lambda_{2k}(A)
        \approx 278.59$ and $\lambda_1(\mathcal{L})\approx 0.2$, illustrating the concentration near $(1-2\sprob)(\gamma_1+\gamma_2)\sqrt{k}\approx 278.28$ and $2\sprob=0.2$, respectively (see \cref{cor:eigvals-adjacency,cor:eigvals-concentration-lap-dense}).}\label{fig:eigvals-demonstration-1}
    \end{figure}

    The proof is based on an application of Bernstein's inequality and the union bound, and is given in \cref{subsec:proofs-ssbm}. Having bounded the minimum degree from below with high probability, we can re-state the Laplacian matrix concentration inequality in the form of the following proposition. Note that the result immediately follows from \cref{th:laplacian-matrix-spectral} and \cref{lem:min-degree-dense}.

    \begin{proposition}\label{prop:laplacian-convergence-dense}
        Let $k\geq 2$ and fix $\gamma_1 > \gamma_2>0$, as well as $0<\sprob<1/2$. Let\break $A\sim\signedblock_k(\gamma_1 k^{-1/2}, \gamma_2 k^{-1/2}, \sprob)$ and let $\mathcal{L}$ be its corresponding normalized Laplacian matrix. Then there exists a constant $C=C(\gamma_1,\gamma_2, \sprob)$ such that for $k$ sufficiently large and with probability at least $1-k^{-2}$, it holds
            \begin{align*}
                \op{\mathcal{L}-\ol{\mathcal{L}}} \leq C k^{-1/4}.
            \end{align*}
    \end{proposition}

    As before, by Weyl's eigenvalue inequality and \cref{lem:properties}, we can immediately estimate the concentration of the eigenvalues of $\mathcal{L}$ as follows.

    \begin{corollary}\label{cor:eigvals-concentration-lap-dense}
        Instate the hypotheses of \cref{prop:laplacian-convergence-dense}. Let\break $A\sim\signedblock_k(\gamma_1 k^{-1/2}, \gamma_2 k^{-1/2}, \sprob)$ and let $\mathcal{L}$ be its corresponding normalized Laplacian matrix. Then for $k$ sufficiently large there exists $C>0$, not depending on $k$, such that with probability at least $1-k^{-2}$, the following eigenvalue estimates hold:
            \begin{align*}
                |\lambda_1(\mathcal{L}) - 2\sprob| &\leq C k^{-1/4}\\
                \left|\lambda_2(\mathcal{L}) - \left(1 - (1-2\sprob)\frac{\gamma_1-\gamma_2}{\gamma_1+\gamma_2}\right)\right| &\leq C k^{-1/4}\\
                \left|\lambda_\ell(\mathcal{L}) - 1\right| &\leq C k^{-1/4},\hspace{.25cm}\ell>2.
            \end{align*}
    \end{corollary}

    The proof is similar to that of \cref{cor:eigvals-adjacency} and is omitted. \cref{fig:eigvals-demonstration-1} illustrates the Laplacian eigenvalue histogram of a moderately large random graph sampled from the dense setting. Next, we turn to a consideration of the eigenvector associated to $\lambda_1(\mathcal{L})$. First we note that
        \begin{align*}
            |\lambda_1(\ol{\mathcal{L}}) - \lambda_2(\ol{\mathcal{L}})| &= \left|2\sprob - 1 + (1-2\sprob)\frac{(\gamma_1-\gamma_2)k^{1/2}+o(1)}{(\gamma_1+\gamma_2)k^{1/2}+o(1)} \right|\\
            &= \left|2\sprob - 1 + (1+o(1))(1-2\sprob)\frac{\gamma_1-\gamma_2}{\gamma_1+\gamma_2} \right| = \Omega(1)
        \end{align*} 
    so that, by the Davis-Kahan theorem (see, e.g., \cite[Corollary 3]{yu2015useful}), we have the following corollary to \cref{prop:laplacian-convergence-dense}. Let $\ol{u_1}= \frac{1}{\sqrt{2k}}\begin{pmatrix}\mathbf{1}_k & -\mathbf{1}_k\end{pmatrix}^T$.

    \begin{corollary}\label{cor:davis-kahan-dense}
        Instate the hypotheses of \cref{prop:laplacian-convergence-dense}. Let $u_1\in\mathbb{R}^{2k}$ denote an eigenvector of $\mathcal{L}$ with unit norm corresponding to $\lambda_1(\mathcal{L})$. Then for $k$ sufficiently large there exists $C>0$, not depending on $k$, such that with probability at least $1-k^{-2}$, the eigenvalue $\lambda_1(\mathcal{L})$ is simple, $u_1$ is uniquely determined up to sign, and there is a choice of sign $\tau\in\{\pm 1\}$ such that
            \begin{align}\label{eq:u1-meanu1}
                \two{\tau u_1 - \ol{u_1}}\leq C k^{-1/4}.
            \end{align}
    \end{corollary}

    \begin{remark}\label{rmk:consistency}\normalfont
        We can consider \cref{cor:davis-kahan-dense} in the context of the signature synchronization problem introduced in \cref{subsec:signed-graphs} as follows. The ground-truth or mean signed graph has Laplacian given by $\ol{\mathcal{L}}$ as in \cref{lem:properties}, which defines a weighted complete graph with two communities given by nodes $[k]$ and $[2k]\setminus[k]$, respectively. In the ground truth, edges within communities have positive signs $\signature=1$ and the edges between communities have negative signs $\signature=-1$, and thus $\mathrm{sgn}(\ol{u_1})$ achieves perfect synchronization in accordance with $\frustration_2(\signature)$ via \cref{eq:frustration-index}. Each signed graph  $A\sim\signedblock_k(\gamma_1 k^{-1/2}, \gamma_2 k^{-1/2}, \sprob)$ can thus be considered a random perturbation of both the edge structure (via $\gamma_1,\gamma_2$) and the signature (via $\sprob$) of the ground truth model. Assuming \cref{eq:u1-meanu1} holds, and letting $\mathcal{S} = \left\{i\in [2k]: \mathrm{sgn}\left({\tau u_1(i)}\right) \neq \mathrm{sgn}(\ol{u_1}(i))\right\}$, we note that
            \begin{align*}
                \two{\tau u_1 - \ol{u_1}}^2 &\geq \sum_{i\in\mathcal{S}} |\tau u_1(i) - \ol{u_1}(i)|^2 \geq \frac{1}{2k} \# \mathcal{S},
            \end{align*}
        which implies $ \# \mathcal{S}\leq C k^{1/2}$. Therefore as $k\rightarrow\infty$, we have that $\# \mathcal{S}/ 2k = o(1)$ with high probability. We therefore conclude that $\mathrm{sgn}(u_1)$ defines a weakly consistent estimator for the solution to the synchronization problem in the ground truth model. 
    \end{remark}

    \subsection{The sparse setting}\label{subsec:sharp-setting}
    In this subsection we provide a second theoretical investigation of the SSBM in the sparse setting where $p,q=\Theta(\log{n}/n)$. This terminology refers to the fact that $p=\log{n}/n$ is a sharp threshold for connectivity in the Erd\H{o}s-R\'{e}nyi model, and therefore we do not in general expect concentration of graph Laplacian matrices along the lines of \cref{th:laplacian-matrix-spectral} unless $p, q \gtrsim \log{n}/n$. This is because whenever $\mathcal{L}$ witnesses an isolated node, $\op{\mathcal{L}-\ol{\mathcal{L}}}= o(1)$ cannot occur. We begin with the off-the-shelf treatment of the adjacency matrices, and then consider Laplacian matrices separately.

    \begin{proposition}\label{prop:sharp-convergence-adj}
        Let $k\geq 2$ and fix $\gamma_1 > \gamma_2>1$, as well as $0<\sprob<1/2$. Let\break $A\sim\signedblock_k(\gamma_1 \log{k}/k, \gamma_2 \log{k}/k, \sprob)$. Then there exists $C>0$, not depending on $k$, such that with probability at least $1-k^{-2}$, it holds
            \begin{align*}
                \op{A-\ol{A}} \leq C\sqrt{\log{k}}.
            \end{align*}
    \end{proposition}

    Note that this proposition follows from \cref{th:adjacency-matrix}. We thus have, by Weyl's eigenvalue inequality, the following eigenvalue concentration inequality.

    \begin{corollary}\label{cor:sharp-eigvals-adjacency}
        Instate the hypotheses of \cref{prop:dense-convergence-adj}. Then there exists $C>0$ such that with probability at least $1-k^{-2}$, the following eigenvalue concentrations hold:
            \begin{align*}
                |\lambda_{2k}(A) - (1-2\sprob)(\gamma_1+\gamma_2)\log{k}| &\leq C\sqrt{\log{k}} + o(1)\\
                |\lambda_{2k-1}(A) - (1-2\sprob)(\gamma_1-\gamma_2)\log{k}| &\leq C\sqrt{\log{k}} + o(1)\\
                |\lambda_\ell(A)| &\leq C\sqrt{\log{k}} + o(1),\hspace{.25cm}\ell < 2k-1.
            \end{align*}
    \end{corollary}

    As we have seen, the key to applying \cref{th:laplacian-matrix-spectral} is controlling the minimum degree of a graph sampled $\signedblock_k(\gamma_1 \log{k}/k, \gamma_2 \log{k}/k, \sprob)$. Our approach is a usage of the multiplicative Chernoff bound. We note that a sharper analysis is possible by using a Poisson approximation and corresponding tail bounds, as in \cite[Lemma 16]{deng2021strong}, but we opt for a simpler approach since $o(1)$ convergence in operator norm is enough for our purposes.

    \begin{lemma}\label{lem:min-degree-sharp}
        Let $k\geq 2$ and fix $\gamma_1 > \gamma_2>1$, as well as $0<\sprob<1/2$. Let\break $A\sim\signedblock_k(\gamma_1 \log{k}/k, \gamma_2 \log{k}/k, \sprob)$. Then there exists a constant $C>0$, not depending on $k$, such that with probability at least $1-k^{-C}$, it holds
            \begin{align*}
                d_{\min} \geq C\left((\gamma_1+\gamma_2)\log{k} - \frac{\gamma_1\log{k}}{k}\right).
            \end{align*}
    \end{lemma}

    We include a proof in \cref{subsec:proofs-ssbm} for completeness. We can now proceed with the concentration results for Laplacian matrices drawn from this setting. Note that this result follows immediately from \cref{th:laplacian-matrix-spectral} and \cref{lem:min-degree-sharp}.

    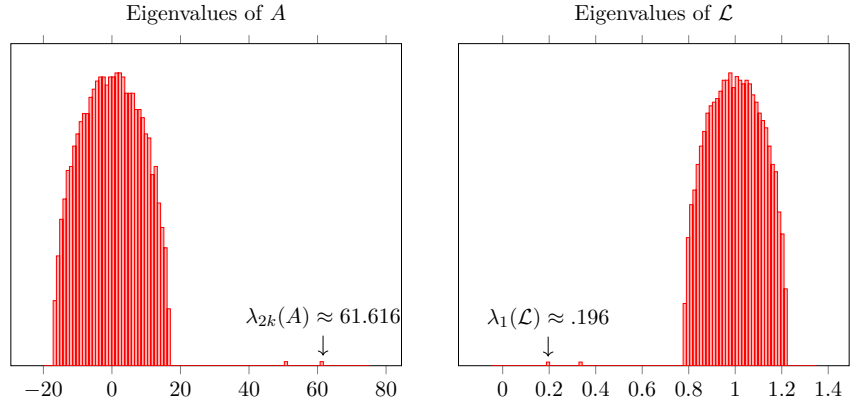
\begin{figure}
        \begin{center}
            \begin{tikzpicture}[scale=.75]
                \begin{axis}[
                    ytick = \empty,
                    name = plot00,
                    ybar,
                    ymin=0,
                    title={Eigenvalues of $A$}
                ]
                \addplot +[
                    hist={
                        bins=100,
                        data min=-20,
                        data max=75
                    },
                    fill=red!30,
                    draw=red
                ] table [y index=0] {eigvals_adj_sparse_01.csv};
                \draw [<-] (axis cs:61.61, 2.5)-- +(0pt,10pt) node[above] {$\lambda_{2k}(A)\approx 61.616$};
                \end{axis}
                \begin{axis}[
                    ytick = \empty,
                    name = plot01,at={(plot00.east)},anchor=west,xshift=1cm,
                    ybar,
                    ymin=0,
                    title={Eigenvalues of $\mathcal{L}$}
                ]
                \addplot +[
                    hist={
                        bins=100,
                        data min=-0.05,
                        data max=1.35
                    },
                    fill=red!30,
                    draw=red
                ] table [y index=0] {eigvals_lap_sparse_01.csv};
                \draw [<-] (axis cs:0.1956, 2.5)-- +(0pt,10pt) node[above] {$\lambda_1(\mathcal{L})\approx .196$};
                \end{axis}
            \end{tikzpicture}
        \end{center}
        \caption{A histogram of the eigenvalues of an observed random signed adjacency matrix $A$ and Laplacian matrix $\mathcal{L}$ obtained from the sparse setting. Here, we take $k=10^3$, $\gamma_1=10$, $\gamma_2=1$, $\sprob=0.1$. In this case, $\lambda_{2k}(A)
        \approx 61.61$ and $\lambda_1(\mathcal{L})\approx 0.19$, illustrating the concentration near $(1-2\sprob)(\gamma_1+\gamma_2)\log{k}\approx 60.78$ and $2\sprob=0.2$, respectively (see \cref{cor:sharp-eigvals-adjacency,cor:eigvals-concentration-lap}).}\label{fig:eigvals-demonstration-2}
    \end{figure}

    \begin{proposition}\label{prop:laplacian-convergence-sharp}
        Let $k\geq 2$ and fix $\gamma_1 > \gamma_2>1$, as well as $0<\sprob<1/2$. Let\break $A\sim\signedblock_k(\gamma_1 \log{k}/k, \gamma_2 \log{k}/k, \sprob)$ and let $\mathcal{L}$ be its associated normalized Laplacian matrix. Then there exist constants $C_1, C_2, C_3$, not depending on $k$, such that with probability at least $1-C_1k^{-C_2}$, it holds
            \begin{align*}
                \op{\mathcal{L}-\ol{\mathcal{L}}} \leq C_3 (\log{k})^{-1/2}
            \end{align*}
    \end{proposition}

    Another application of Weyl's inequality gives us the eigenvalue bounds as well.

    \begin{corollary}\label{cor:eigvals-concentration-lap}
        Instate the hypotheses of \cref{prop:laplacian-convergence-sharp}. Then there exist constants $C_1, C_2, C_3>0$, not depending on $k$, such that with probability at least $1-C_1n^{-C_2}$, the following eigenvalue estimates hold:
            \begin{align*}
                |\lambda_1(\mathcal{L}) - 2\sprob| &\leq C_3 (\log{k})^{-1/2}\\
                \left|\lambda_2(\mathcal{L}) - \left(1 - (1-2\sprob)\frac{\gamma_1-\gamma_2}{\gamma_1+\gamma_2}\right)\right| &= o(1)\\
                \left|\lambda_\ell(\mathcal{L}) - 1\right| &=o(1),\hspace{.25cm}\ell>2.
            \end{align*}
    \end{corollary}

    \cref{fig:eigvals-demonstration-2} illustrates a histogram of the Laplacian eigenvalues of a random graph sampled from the sparse regime under consideration. Finally, we can estimate the concentration of the leading eigenvector of $\mathcal{L}$ near that of $\ol{\mathcal{L}}$.

    \begin{corollary}\label{cor:davis-kahan-sharp}
        Instate the hypotheses of \cref{prop:laplacian-convergence-sharp}. Let $u_1\in\mathbb{R}^{2k}$ denote an eigenvector of $\mathcal{L}$ with unit norm corresponding to $\lambda_1(\mathcal{L})$. Then for $k$ sufficiently large there exist constants $C_1, C_2>0$, not depending on $k$, such that with probability at least $1-k^{-C_1}$, the eigenvalue $\lambda_1(\mathcal{L})$ is simple, $u_1$ is uniquely determined up to sign, and there is a choice of sign $\tau\in\{\pm 1\}$ such that
            \begin{align}\label{eq:u1-meanu1-sharp}
                \two{\tau u_1 - \ol{u_1}}\leq C_2(\log{k})^{-1/2}.
            \end{align}
    \end{corollary}

    \section{Experimental data}\label{sec:experimental-data}

    In this section we include some experimental observations intended to supplement the theoretical contributions of the paper.

    \begin{figure}[h!]
        \begin{center}
            \includegraphics[width=0.8\textwidth]{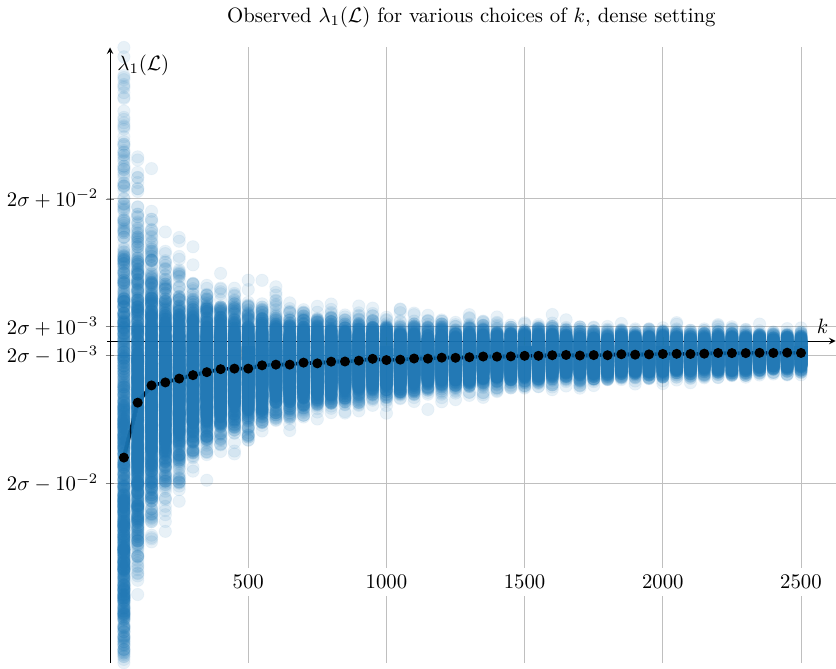}

            \includegraphics[width=0.8\textwidth]{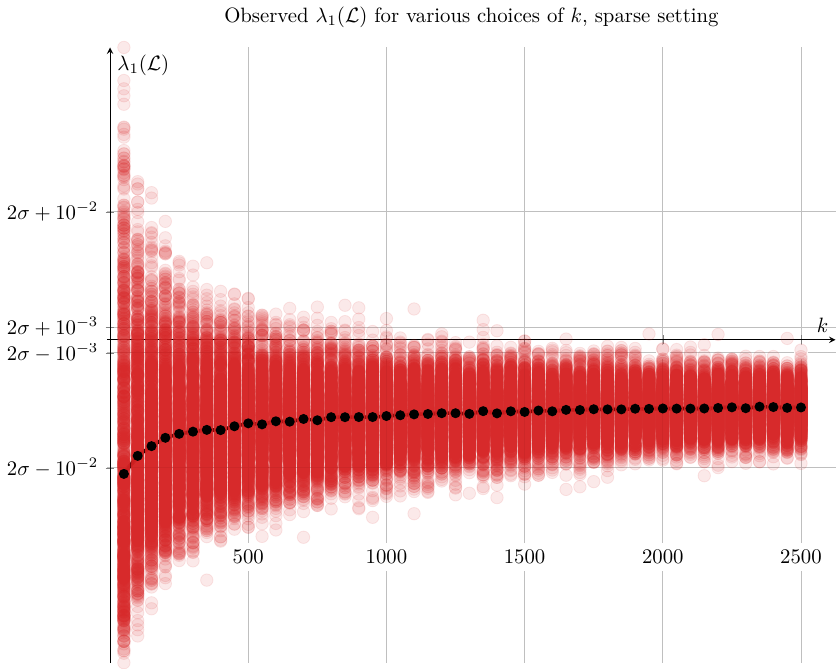}
        \end{center}
        \caption{We fix $\gamma_1 = 10$, $\gamma_2=1$, $\sprob=0.1$ and investigate the concentration of $\lambda_1(\mathcal{L})$ near $2\sprob$ for values of $k$ in the range $k=50, 100, \dotsc, 2500$ (see~\cref{cor:eigvals-concentration-lap} and~\cref{cor:eigvals-concentration-lap-dense}). For each $k$ we sampled $1000$ observations from $\signedblock_k$ in each of the dense and sparse settings, and then recorded the spectral gap $\lambda_1(\mathcal{L})$. \textbf{(Above)} Data obtained from the dense setting $\signedblock_k(\gamma_1k^{-1/2}, \gamma_2k^{-1/2}, 0.1)$. Mean eigenvalues are indicated by black dots. \textbf{(Below)} Data obtained from the sparse setting $\signedblock_k(\gamma_1\log{k}/k, \gamma_2\log{k}/k, 0.1)$.}
    \end{figure}

    \begin{figure}[h!]
        \begin{center}
            \includegraphics[width=0.75\textwidth]{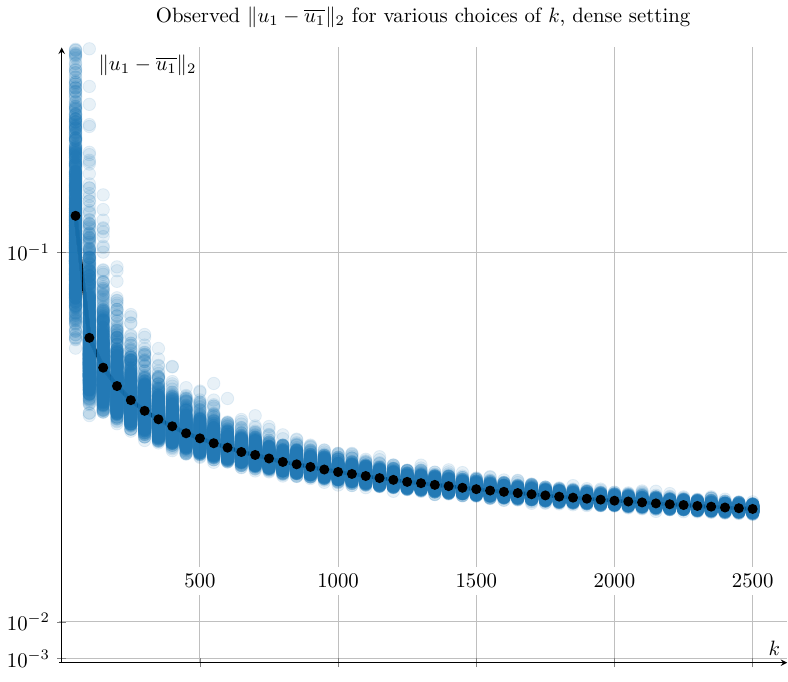}

            \includegraphics[width=0.75\textwidth]{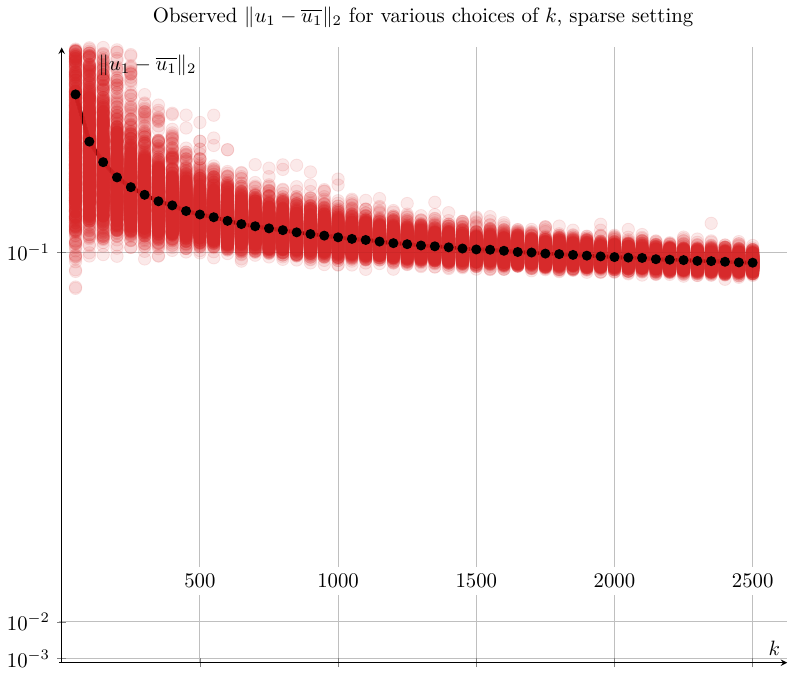}
        \end{center}
        \caption{We fix $\gamma_1 = 10$, $\gamma_2=1$, $\sprob=0.1$ and investigate the concentration of $\two{u_1-\ol{u_1}}$ for values of $k$ in the range $k=50, 100, \dotsc, 2500$ (see~\cref{cor:davis-kahan-dense} and~\cref{cor:davis-kahan-sharp}). For each $k$ we sampled $1000$ observations from $\signedblock_k$ in each of the dense and sparse settings, and then recorded $\min_{\tau\in\{\pm1\}}\two{\tau u_1-\ol{u_1}}$. \textbf{(Above)} Data obtained from the dense setting $\signedblock_k(\gamma_1k^{-1/2}, \gamma_2k^{-1/2}, 0.1)$. Mean values of $\two{u_1-\ol{u_1}}$ are indicated by black dots. \textbf{(Below)} Data obtained from the sparse setting $\signedblock_k(\gamma_1\log{k}/k, \gamma_2\log{k}/k, 0.1)$.}
    \end{figure}

    \newpage

    \begin{figure}[h!]
        \begin{center}
            \includegraphics[width=0.8\textwidth]{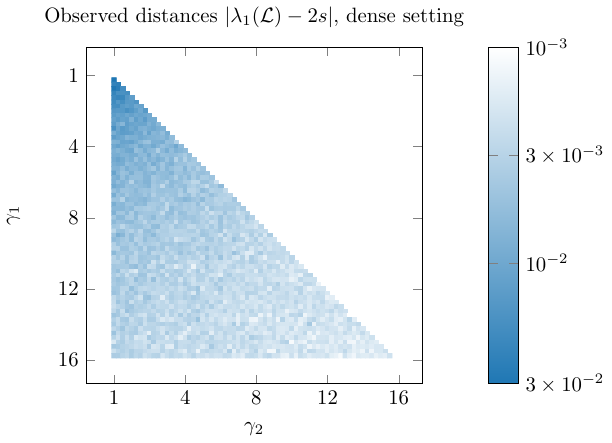}

            \includegraphics[width=0.8\textwidth]{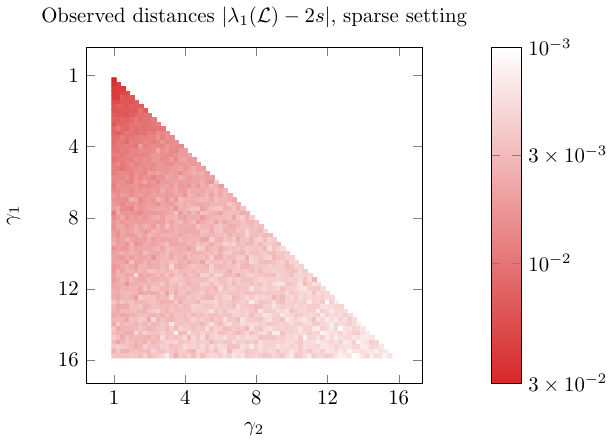}
        \end{center}
        \caption{We fix $k=256$, $\sprob=0.1$ and investigate the concentration of $|\lambda_1(\mathcal{L}) - 2\sprob|$ for values of $\gamma_1$ and $\gamma_2$ (see~\cref{cor:eigvals-concentration-lap} and~\cref{cor:eigvals-concentration-lap-dense}). For each pair $\gamma_1,\gamma_2$ satisfying $\gamma_1>\gamma_2$ in the range $\gamma_i = 1, 1.23, \dotsc, 16$ (with $64$ steps), we sampled $25$ observations from $\signedblock_{k=256}$ in each of the dense and sparse settings, and then recorded the mean of $|\lambda_1(\mathcal{L}) - 2\sprob|$ across the 25 observations. \textbf{(Above)} Data obtained from the dense setting $\signedblock_{k=256}(\gamma_1k^{-1/2}, \gamma_2k^{-1/2}, 0.1)$. \textbf{(Below)} Data obtained from the sparse setting $\signedblock_{k=256}(\gamma_1\log{k}/k, \gamma_2\log{k}/k, 0.1)$. }
    \end{figure}

    \newpage

    \begin{figure}[h!]
        \begin{center}
            \includegraphics[width=0.8\textwidth]{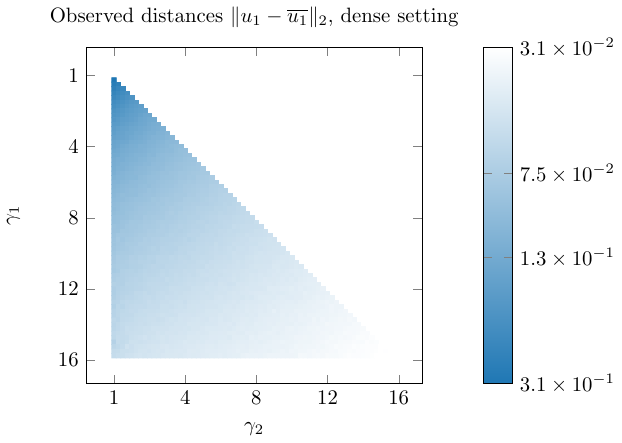}

            \includegraphics[width=0.8\textwidth]{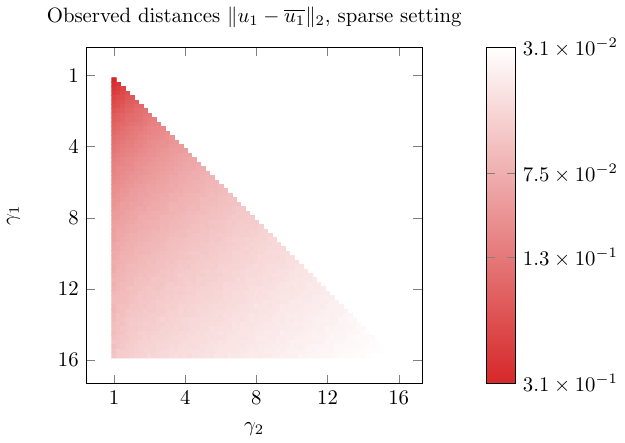}
        \end{center}
        \caption{We fix $k=256$, $\sprob=0.1$ and investigate the concentration of $\two{u_1-\ol{u_1}}$ for values of $\gamma_1$ and $\gamma_2$ (see~\cref{cor:davis-kahan-dense} and~\cref{cor:davis-kahan-sharp}). For each pair $\gamma_1,\gamma_2$ satisfying $\gamma_1>\gamma_2$ in the range $\gamma_i = 1, 1.23, \dotsc, 16$ (with $64$ steps), we sampled $25$ observations from $\signedblock_{k=256}$ in each of the dense and sparse settings, and then recorded the mean of $\min_{\tau\in\{\pm1\}}\two{\tau u_1-\ol{u_1}}$ across the 25 observations. \textbf{(Above)} Data obtained from the dense setting $\signedblock_{k=256}(\gamma_1k^{-1/2}, \gamma_2k^{-1/2}, 0.1)$. \textbf{(Below)} Data obtained from the sparse setting $\signedblock_{k=256}(\gamma_1\log{k}/k, \gamma_2\log{k}/k, 0.1)$. }
    \end{figure}

    \newpage

    \begin{figure}[h!]
        \begin{center}
            \includegraphics[width=\textwidth]{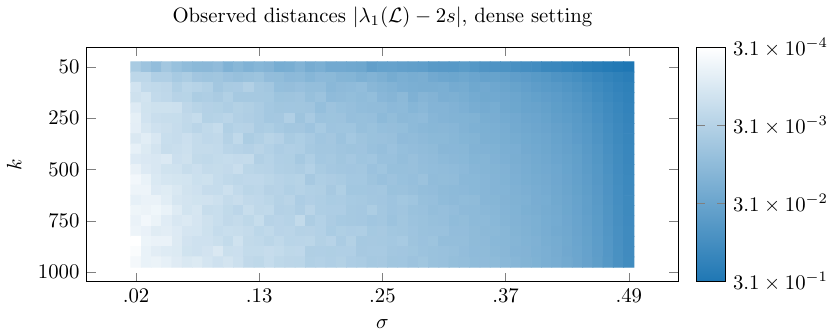}

            \includegraphics[width=\textwidth]{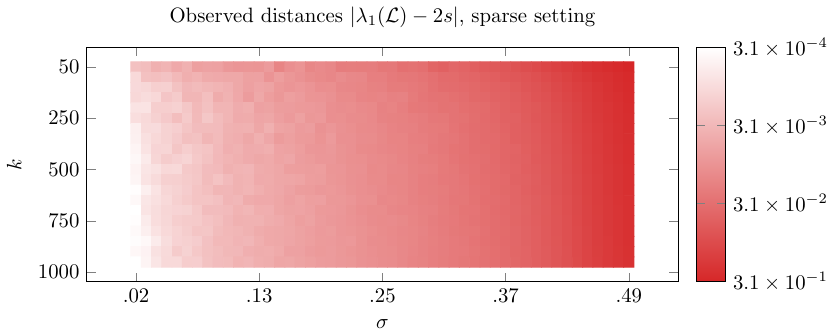}
        \end{center}
        \caption{We fix $\gamma_1=10, \gamma_2=1$ and investigate the concentration of $|\lambda_1(\mathcal{L}-2\sprob)|$ for values of $k$ and $\sprob$ (see~\cref{cor:eigvals-concentration-lap} and~\cref{cor:eigvals-concentration-lap-dense}). For each $k$ in the range $k=5, 100, \dotsc, 1000$ and $\sprob$ in the range $\sprob=0.02, 0.029,\dotsc, 0.49$ (with $50$ steps), we sampled $25$ observations from $\signedblock_{k}$ in each of the dense and sparse settings, and then recorded the mean of $|\lambda_1(\mathcal{L}) - 2\sprob|$ across the 25 observations. \textbf{(Above)} Data obtained from the dense setting $\signedblock_{k}(\gamma_1k^{-1/2}, \gamma_2k^{-1/2}, \sprob)$. \textbf{(Below)} Data obtained from the sparse setting $\signedblock_{k}(\gamma_1\log{k}/k, \gamma_2\log{k}/k, \sprob)$.}
    \end{figure}

    \newpage

    \begin{figure}[h!]
        \begin{center}
            \includegraphics[width=\textwidth]{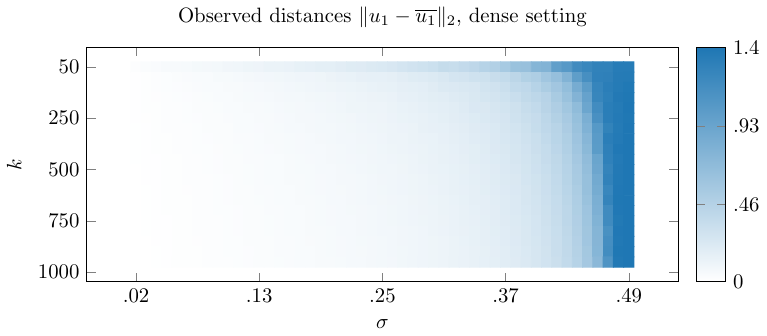}

            \includegraphics[width=\textwidth]{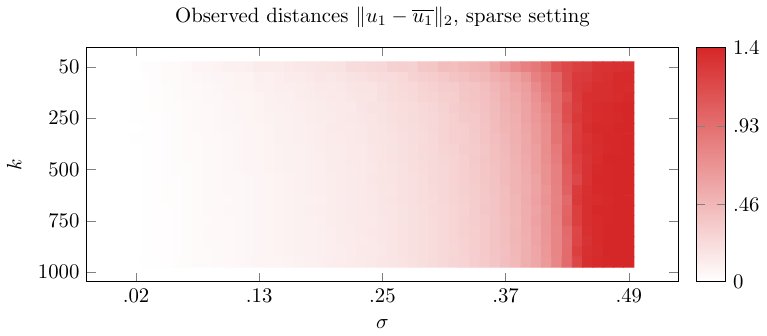}
        \end{center}
        \caption{We fix $\gamma_1=10, \gamma_2=1$ and investigate the concentration of $\two{u_1-\ol{u_1}}$ for values of $k$ and $\sprob$ (see~\cref{cor:davis-kahan-dense} and~\cref{cor:davis-kahan-sharp}). For each $k$ in the range $k=5, 100, \dotsc, 1000$ and $\sprob$ in the range $\sprob=0.02, 0.029,\dotsc, 0.49$ (with $50$ steps), we sampled $25$ observations from $\signedblock_{k}$ in each of the dense and sparse settings, and then recorded the mean of $\min_{\tau\in\{\pm1\}}\two{\tau u_1-\ol{u_1}}$ across the 25 observations. \textbf{(Above)} Data obtained from the dense setting $\signedblock_{k}(\gamma_1k^{-1/2}, \gamma_2k^{-1/2}, \sprob)$. \textbf{(Below)} Data obtained from the sparse setting $\signedblock_{k}(\gamma_1\log{k}/k, \gamma_2\log{k}/k, \sprob)$.}
    \end{figure}

    \section*{Acknowledgements}

    The author wishes to acknowledge financial support from the Halicio{\u g}lu Data Science Institute at UC San Diego in the form of their Graduate Prize Fellowship. The author has no competing interests or conflicts of interest to disclose.

    \appendix

    \section{Missing proofs}\label{sec:missing-proofs}

    This appendix contains proofs of various results which appear throughout the paper.

    \subsection{Proofs from \cref{sec:background}}\label{subsec:proofs-background}

    \begin{proof}[Proof of \cref{lem:properties}]
        Statement \textit{(i)} is straightforward to see based on the definition of $\signedblock_k(p, q, \sprob)$ and the fact that its underlying graph follows a stochastic bisection model where, for each node $i\in[2k]$, $k-1$ edges may occur independently, each with probability $p$, and $k$ additional edges may occur independently, each with probability $q$. 

        The first half of statement \textit{(ii)} follows along similar lines. For the second half, let $\theta_1 = p(1-2\sprob)$ and $\theta_2 = q(2\sprob-1)$. Then
            \begin{align*}
                \ol{A} &= \begin{pmatrix}
                    \theta_1\widetilde{J}_{k\times k} & \theta_2{J}_{k\times k}\\
                    \theta_2{J}_{k\times k} & \theta_1\widetilde{J}_{k\times k}
                \end{pmatrix}=-\theta_1 I_{n\times n} + \theta_1\begin{pmatrix}
                    J_{k\times k} & \\
                    & J_{k\times k}
                \end{pmatrix} + \theta_2\begin{pmatrix}
                    & J_{k\times k} \\
                    J_{k\times k} & 
                \end{pmatrix}
            \end{align*}
        Thus, $\ol{A}$ is a rank 2 perturbation of the identity, and moreover, we have the simultaneous diagonalization
            \begin{align*}
                \begin{pmatrix}
                J_{k\times k} & \\
                & J_{k\times k}
                \end{pmatrix} \begin{pmatrix}
                    \mathbf{1}_k\\
                    \mathbf{1}_k
                \end{pmatrix} &= k \begin{pmatrix}
                    \mathbf{1}_k\\
                    \mathbf{1}_k
                \end{pmatrix},\hspace{.5cm}\begin{pmatrix}
                    J_{k\times k} & \\
                    & J_{k\times k}
                    \end{pmatrix} \begin{pmatrix}
                        \mathbf{1}_k\\
                        -\mathbf{1}_k
                    \end{pmatrix} = k \begin{pmatrix}
                        \mathbf{1}_k\\
                        -\mathbf{1}_k
                    \end{pmatrix}\\
                \begin{pmatrix}
                     & J_{k\times k} \\
                    J_{k\times k} & 
                    \end{pmatrix} \begin{pmatrix}
                        \mathbf{1}_k\\
                        \mathbf{1}_k
                    \end{pmatrix} &= k \begin{pmatrix}
                        \mathbf{1}_k\\
                        \mathbf{1}_k
                    \end{pmatrix},\hspace{.5cm}\begin{pmatrix}
                        & J_{k\times k} \\
                       J_{k\times k} & 
                       \end{pmatrix} \begin{pmatrix}
                            \mathbf{1}_k\\
                            -\mathbf{1}_k
                        \end{pmatrix} = -k \begin{pmatrix}
                            \mathbf{1}_k\\
                            -\mathbf{1}_k
                        \end{pmatrix}.
            \end{align*}
        Here, $\mathbf{1}_k = J_{k\times 1}\in\mathbb{R}^k$ is the vector of all ones. Thus $\ol{A}$ has eigenvalues $-\theta_1 + \theta_1k + \theta_2k, -\theta_1+ \theta_1k - \theta_2k$, each with multiplicity 1, corresponding to the nontrivial contributions from the perturbation matrices diagonalized above. The rest of the eigenvalues are copies of $-\theta_1$. We have, for example,
            \begin{align*}
                (k-1)\theta_1 + \theta_2 &= (k-1)p(1-2\sprob)+ qk(2\sprob-1)\\
                &= (1-2\sprob)(p(k-1) -qk)
            \end{align*}
        and the rest follow similarly.

        Statement \textit{(iii)} follows along the same lines, with the initial observation being that $\ol{D}^{-1/2}\ol{A}\ol{D}^{-1/2} = \ol{d}^{-1}\ol{A}$, and thus the eigenvalues of $\ol{\mathcal{L}}$ occur as $1-\ol{d}^{-1}\lambda$ where $\lambda$ is any eigenvalue of $\ol{A}$.
    \end{proof}

    \subsection{Proofs from \cref{sec:concentration-results}}\label{subsec:proofs-concentration}

    \begin{proof}[Proof of \cref{lemma:light-pairs}]
        Let $t>0$ be fixed. By the union bound and the estimate $\card{\mathfrak{B}}\leq e^{2n\log{9}/\delta}$,
            \begin{align*}
                \pr{\sup_{x, y\in \mathfrak{B}} \left|\sum_{(i, j)\in\mathfrak{L}(x, y)} x_i w_{ij} y_j\right| \geq t\sqrt{\alpha}} &\leq \sum_{x, y\in \mathfrak{B}} \pr{ \left|\sum_{(i, j)\in\mathfrak{L}(x, y)} x_i w_{ij} y_j\right| \geq t\sqrt{\alpha}}\\
                &\leq e^{2n\log(9/\delta) }\pr{ \left|\sum_{(i, j)\in\mathfrak{L}(x, y)} x_i w_{ij} y_j\right| \geq t\sqrt{\alpha}}.
            \end{align*}
        Letting $x, y\in\mathfrak{B}$ be fixed, we define $b_{ij} = x_i y_j \mathbf{1}(|x_i y_j| \leq \sqrt{\alpha} / n) + x_j y_i \mathbf{1}(|x_i y_j| \leq \sqrt{\alpha}/n) $. Then we have
            \begin{align*}
                \sum_{(i, j)\in\mathfrak{L}(x, y)} x_i w_{ij} y_j = \sum_{1\leq i \leq j \leq n}b_{ij}w_{ij}.
            \end{align*}
        Now since $|b_{ij}|\leq 2\sqrt{\alpha}/n$ and $|w_{ij}|\leq 2$, it follows that each term in the sum is mean zero and is bounded above by $ 4\sqrt{\alpha}/n$ almost surely. Set $\nu^2 := \sum_{i\leq j} \ep{b_{ij}^2w_{ij}^2}$. By Bernstein's inequality,
            \begin{align*}
                \pr{\left| \sum_{1\leq i \leq j \leq n}b_{ij}w_{ij} \right| \geq t\sqrt{\alpha}} &\leq 2 \exp\left\{ -\frac{\frac{1}{2}t^2d}{\nu^2 + \frac{t}{3}\frac{4 d}{n}}\right\} \\
                &= 2 \exp\left\{ -\frac{nt^2 }{\frac{2n}{{\alpha}}\nu^2 + \frac{8t}{3}}\right\}.
            \end{align*}
        Since
            \begin{align*}
                \sum_{i\leq j}b_{ij}^2 \leq 2\sum_{i, j=1}^n |x_i y_j|^2 \leq 2\two{x}^2\two{y}^2 = 2,
            \end{align*}
        it follows that $\nu^2 \leq 2\max_{i, j}\ep{w_{ij}^2}$, and in turn,
            \begin{align*}
                2 \exp\left\{ -\frac{nt^2 }{\frac{2n}{{\alpha}}\nu^2 + \frac{8t}{3}}\right\} &\leq 2 \exp\left\{ -\frac{nt^2 }{\frac{4n}{{\alpha}}\max_{i, j}\ep{w_{ij}^2} + \frac{8t}{3}}\right\}
            \end{align*}
        Next, focusing on $\max_{i,j} \ep{w_{ij}^2} = \max_{i,j}\va{a_{ij}}$, we recall that $a_{ij} = P_{ij}(1-2S_{ij})$ and that $P_{ij}, S_{ij}$ are independent. Therefore, we have
            \begin{align*}
                \va{ P_{ij}(1-2S_{ij})} &= \ep{P_{ij}^2}\ep{(1-2S_{ij})^2} - \ep{P_{ij}}^2\ep{1 - 2S_{ij}}^2\\
                &= \eprob_{ij} - \eprob_{ij}^2(1-2\sprob_{ij})^2\\
                &=\frac{|\ol{a}_{ij}|}{|1 - 2\sprob_{ij}|} - \ol{a}_{ij}^2
            \end{align*}
        Let $\rho = \min_{i, j} | \sprob_{ij} -1/2|$. It holds $|1 - 2\sprob_{ij}|\geq 2\rho$, and thus
            \begin{align*}
                \frac{|\ol{a}_{ij}|}{|1 - 2\sprob_{ij}|} - \ol{a}_{ij}^2\leq \frac{|\ol{a}_{ij}|}{2\rho} - \ol{a}_{ij}^2 \leq |\ol{a}_{ij}|\left(\frac{1}{2\rho} - |\ol{a}_{ij}|\right)\leq \frac{1}{2\rho}|\ol{a}_{ij}|. 
            \end{align*} 
        Therefore it follows that $\max_{i, j} \ep{w_{ij}^2} \leq |\ol{a}|_{\max}/(2c_0)$, where $|\ol{a}|_{\max} = \max_{i, j}|\ol{a}_{ij}|$. In turn, continuing from before, we have
            \begin{align*}
                \pr{\left| \sum_{1\leq i \leq j \leq n}b_{ij}w_{ij} \right| \geq t\sqrt{\alpha}} &\leq 2 \exp\left\{ -\frac{nt^2 }{\frac{4n}{{2c_0 d}}|\ol{a}|_{\max} + \frac{8t}{3}}\right\}\\
                &\leq 2 \exp\left\{ -\frac{nt^2 }{\frac{2}{{c_0}} + \frac{8t}{3}}\right\}
            \end{align*}
        The claim follows.
    \end{proof}


    \begin{proof}[Proof of \cref{lemma:heavy-pairs}]
        We begin by noting that for each $x,y\in \mathfrak{B}$,
            \begin{align*}
                \left|\sum_{(i, j)\in\ol{\mathfrak{L}}(x, y)} x_i y_j \ol{a}_{ij}\right| & \leq \sum_{(i, j)\in\ol{\mathfrak{L}}(x, y)} \left|\frac{x_i^2 y_j^2}{x_i y_j}\ol{a}_{ij}\right|\leq |\ol{a}|_{\max} \frac{n}{\sqrt{\alpha}} \sum_{i, j=1}^n x_i^2 y_j^2\leq \sqrt{\alpha}
            \end{align*}
        since $n|\ol{a}|_{\max}\leq \alpha$ and $x, y\in B$. Therefore it is enough to show that $\left|\sum_{(i, j)\in\ol{\mathfrak{L}}(x, y)} x_i y_j a_{ij}\right| \leq C\sqrt{\alpha}$ with high probability. For this task we employ a small sleight of hand. We begin by dividing up the indices into four sets based on the signs of the entries, namely, let $\mathfrak{C}_{++} :=\ol{\mathfrak{L}}(x, y)\cap \{(i, j) : x_i > 0, y_i >0 \}$, $\mathfrak{C}_{-+} :=  \ol{\mathfrak{L}}(x, y)\cap \{(i, j) : x_i < 0, y_i >0 \}$, and similarly for $\mathfrak{C}_{+-}$ and $\mathfrak{C}_{--}$. Then, we have that
            \begin{align*}
                \left|\sum_{(i, j)\in\ol{\mathfrak{L}}(x, y)} x_i y_j a_{ij}\right| & \leq \left|\sum_{(i, j) \in \mathfrak{C}_{++}} x_i y_j a_{ij}\right| + \left|\sum_{(i, j) \in \mathfrak{C}_{-+}} x_i y_j a_{ij}\right| \\
                &+ \left|\sum_{(i, j) \in \mathfrak{C}_{+-}} x_i y_j a_{ij}\right| + \left|\sum_{(i, j) \in \mathfrak{C}_{--}} x_i y_j a_{ij}\right|.
            \end{align*}
        Thus it suffices to show that each term on the right-hand side is of order $O(\sqrt{\alpha})$ with high probability. Now, we observe that
            \begin{align*}
                \left|\sum_{(i, j) \in \mathfrak{C}_{++}} x_i y_j a_{ij}\right| \leq \left|\sum_{(i, j) \in \mathfrak{C}_{++}} x_i y_j |a_{ij}|\right|
            \end{align*}
        since all of the signs of the products $x_iy_j$ agree, and thus the inclusion of some negative values of $a_{ij}$ serves to shrink the overall absolute value of the sum. But the matrix $|A| = (|a_{ij}|)_{i, j=1}^n$ is a random unsigned adjacency matrix satisfying the conditions required for concentration of measure in the unsigned setting. Therefore, by \cite[Lemma 4.3]{supplei2015consistency}, we have that for any $r>0$, there exists $C>0$ such that with probability $1 - n^{-r}$, it holds 
            \begin{align*}
                \sup_{x, y\in\mathfrak{B}}\left|\sum_{(i, j) \in \mathfrak{C}_{++}} x_i y_j |a_{ij}|\right| \leq C\sqrt{\alpha}.
            \end{align*}
        Note that $C$ depends on \textit{both} $r$ and $c_0$, since $\ol{a}_{ij}\leq\alpha/n$ guarantees only that $\eprob_{ij}\leq \alpha / (2c_0 n)$. The claim follows.
    \end{proof}


    \begin{proof}[Proof of \cref{th:laplacian-matrix-spectral}]
        We have the initial estimate
            \begin{align}
                \op{\mathcal{L} - \ol{\mathcal{L}}} &= \op{D^{-1/2}AD^{-1/2} - \ol{D}^{-1/2}\ol{A}\ol{D}^{-1/2}} \\
                &\leq \op{D^{-1/2}(A - \ol{A})D^{-1/2}} + \op{D^{-1/2}\ol{A}D^{-1/2} -\ol{D}^{-1/2}\ol{A}\ol{D}^{-1/2}}
            \end{align}
        The first term can be controlled using \cref{th:adjacency-matrix}, namely,
            \begin{align}
                \op{D^{-1/2}(A - \ol{A})D^{-1/2}} \leq \frac{\op{A - \ol{A}}}{d_{\min}}\leq C_1(c_0, c_1, r)\frac{\sqrt{\alpha}}{d_{\min}}
            \end{align}
        with probability $1 - n^{-r}$. For the second term, let $d = |A|\mathbf{1}_n$ and $\ol{d} = \eprmatrix\mathbf{1}_n$. Then,
            \begingroup
            \allowdisplaybreaks
            \begin{align*}
                &\op{D^{-1/2}\ol{A}D^{-1/2} -\ol{D}^{-1/2}\ol{A}\ol{D}^{-1/2}} \\
                &\leq \fro{D^{-1/2}\ol{A}D^{-1/2} -\ol{D}^{-1/2}\ol{A}\ol{D}^{-1/2}} \\
                &\leq \sqrt{\sum_{i, j=1}^n \ol{a}_{ij}^2 \left(\frac{1}{\sqrt{d_i d_j}} - \frac{1}{\sqrt{\ol{d}_i \ol{d}_j}} \right)^2  } \\
                &\leq |\ol{a}|_{\max} \sqrt{\sum_{i, j=1}^n \left(\frac{\sqrt{\ol{d}_i \ol{d}_j} - \sqrt{d_i d_j} }{\sqrt{d_i d_j\ol{d}_i\ol{d}_j}} \right)^2  } \\
                &\leq  |\ol{a}|_{\max} \sqrt{\sum_{i, j=1}^n \left(\frac{\ol{d}_i \ol{d}_j - d_i d_j }{\left(   \sqrt{\ol{d}_i \ol{d}_j} + \sqrt{d_i d_j}       \right)\sqrt{d_i d_j\ol{d}_i\ol{d}_j}} \right)^2  } \\
                &\leq \frac{|\ol{a}|_{\max}}{d_{\min} \ol{d}_{\min}} \sqrt{\sum_{i, j=1}^n \left(\frac{\ol{d}_i \ol{d}_j - d_i d_j }{\sqrt{\ol{d}_i \ol{d}_j} + \sqrt{d_i d_j}} \right)^2  }\\
                &\leq \frac{|\ol{a}|_{\max}}{2 \min\left\{  d_{\min}, \ol{d}_{\min} \right\}^3} \sqrt{\sum_{i, j=1}^n \left( {\ol{d}_i \ol{d}_j} - {d_i d_j} \right)^2  }\\
                &\leq \frac{|\ol{a}|_{\max}}{2 \min\left\{  d_{\min}, \ol{d}_{\min} \right\}^3} \fro{dd^T - \ol{d}\ol{d}^T}\\
                &\leq \frac{|\ol{a}|_{\max}}{2 \min\left\{  d_{\min}, \ol{d}_{\min} \right\}^3} \left(\fro{d(d - \ol{d})^T } + \fro{\ol{d}(d-\ol{d})^T}\right)\\
                &\leq \frac{|\ol{a}|_{\max}}{2 \min\left\{  d_{\min}, \ol{d}_{\min} \right\}^3}  \left(\two{d}\two{d - \ol{d}} +\two{\ol{d}}\two{d-\ol{d}}\right)\\
                &\leq \frac{|\ol{a}|_{\max}}{2 \min\left\{  d_{\min}, \ol{d}_{\min} \right\}^3}  \left(\op{|A|} + \op{\eprmatrix} \right)\op{|A|- \eprmatrix}\two{\mathbf{1}_n}^2\\
                &\leq \frac{|\ol{a}|_{\max}}{2 \min\left\{  d_{\min}, \ol{d}_{\min} \right\}^3}  \left(\op{|A| - \eprmatrix} + 2\op{\eprmatrix} \right)\op{|A|- \eprmatrix}\two{\mathbf{1}_n}^2
            \end{align*}
            \endgroup
        Now we make use of \cref{thm:unsigned-adjacency} and the assumptions directly made on $\eprob_{ij}$ to bound $\op{|A|-\eprmatrix}$, i.e., the fact that the underlying unsigned adjacency matrix is also concentrations sufficiently rapidly. Note, as in the proof of \cref{lemma:heavy-pairs}, the corresponding constants will depend on $c_0$ in addition to $r$. This coupled with with the additional fact that $|\ol{a}|_{\max} \leq \alpha/n$, we have that
            \begin{align*}
                &\frac{|\ol{a}|_{\max}}{2 \min\left\{  d_{\min}, \ol{d}_{\min} \right\}^3}  \left(\op{|A| - \eprmatrix} + 2\op{\eprmatrix} \right)\op{|A| - \eprmatrix}\two{\mathbf{1}_n}^2\\
                &\leq C_2(c_0, c_1, r) \frac{|\ol{a}|_{\max}}{2 \min\left\{  d_{\min}, \ol{d}_{\min} \right\}^3}\left(\sqrt{\alpha}  +  \alpha\right)n \sqrt{\alpha}\\
                &\leq C_3(c_0, c_1, r) \frac{\alpha ^{5/2}}{ \min\left\{  d_{\min}, \ol{d}_{\min} \right\}^3}
            \end{align*}
        with probability at least $1 - 2n^{-r}$. Since $\op{\ol{A}} \leq \fro{\ol{A}} \leq \alpha$, upon combining the two estimates, we have
            \begin{align*}
                \op{\mathcal{L} - \ol{\mathcal{L}}} &\leq  C_1(c_0, c_1, r)\frac{\alpha^{1/2}}{d_{\min}} + C_3(c_0, c_1, r)\frac{\alpha ^{5/2}}{ \min\left\{  d_{\min}, \ol{d}_{\min} \right\}^3}\\
                &\leq  C_1(c_0, c_1, r)\frac{\ol{d}_{\min}^2 \alpha^{1/2}}{\ol{d}_{\min}^2d_{\min}} +  C_3(c_0, c_1, r) \frac{\alpha^{5/2}}{ \min\left\{  d_{\min}, \ol{d}_{\min} \right\}^3}\\
                &\leq C_4(c_0, c_1, r)\frac{\alpha^{5/2}}{ \min\left\{  d_{\min}, \ol{d}_{\min} \right\}^3}
            \end{align*}
        with probability at least $1 - 3n^{-r}$, as claimed.
    \end{proof}

    \subsection{Proofs from \cref{sec:ssbm}}\label{subsec:proofs-ssbm}

    \begin{proof}[Proof of \cref{cor:eigvals-adjacency}]
        From \cref{lem:properties}, we have that the leading eigenvalue of $\ol{A}$ is given by
            \begin{align*}
                \lambda_{2k}(A) &= (1-2\sprob)(\gamma_1k^{1/2} + \gamma_2 k^{1/2} - \gamma_1k^{-1/2})
            \end{align*}
        Therefore we have
            \begin{align*}
                |\lambda_{2k}(A) - (1-2\sprob)(\gamma_1k^{1/2} + \gamma_2 k^{1/2} - \gamma_1k^{-1/2})| \geq |\lambda_{2k}(A) - (1-2\sprob)(\gamma_1 + \gamma_2)k^{1/2}| + o(1)
            \end{align*}
        By Weyl's eigenvalue inequality, it holds
            \begin{align*}
                |\lambda_{2k}(A) - (1-2\sprob)(\gamma_1 + \gamma_2)k^{1/2}| \leq \op{A-\ol{A}} + o(1),
            \end{align*}
        and thus the first eigenvalue concentration claim follows from \cref{prop:dense-convergence-adj}. The rest are similar.
    \end{proof}


    \begin{proof}[Proof of \cref{lem:min-degree-dense}]
        We begin by fixing $i$, which without loss of generality we take to be $i=1$, and observing that
            \begin{align*}
                d_1 = \sum_{j=2}^{k}\mathbf{1}_{j\sim i} + \sum_{j=k+1}^{2k}\mathbf{1}_{j\sim i}.
            \end{align*}
        Here, $\mathbf{1}_{j\sim i}$ denotes the indicator random variable for the event $\{i, j\}\in E$; thus for $2\leq j \leq k$, we have $\mathbf{1}_{j\sim i}\sim\mathsf{Bernoulli}(\gamma_1 k^{-1/2})$, and for $k+1\leq j \leq 2k$, we have $\mathbf{1}_{j\sim i}\sim\mathsf{Bernoulli}(\gamma_2 k^{-1/2})$. Therefore we have
            \begin{align*}
                \va{\mathbf{1}_{j\sim i}} &= \gamma_1 k^{-1/2}(1-\gamma_1 k^{-1/2}),\hspace{.25cm}2\leq j \leq k\\
                \va{\mathbf{1}_{j\sim i}} &= \gamma_2 k^{-1/2}(1-\gamma_2 k^{-1/2}),\hspace{.25cm}k+1\leq j \leq 2k.
            \end{align*}
        In either case, $\va{\mathbf{1}_{j\sim i}}\leq \gamma_1k^{-1/2}$. Therefore by Bernstein's inequality we have that for each $t>0$, it holds
            \begin{align*}
                \pr{|d_1 - \ol{d}| \geq t} \leq\exp\left\{-\frac{\frac{1}{2}t^2}{2\gamma_1k^{1/2} + \frac{1}{3}t}\right\}.
            \end{align*}
        In particular taking $t=\gamma_1 k^{1/3}$, we have that
            \begin{align*}
                \pr{|d_1 - \ol{d}| \geq \gamma_1 k^{1/3}} &\leq \exp\left\{-\frac{\frac{1}{2}\gamma_1^2 k^{2/3}}{2\gamma_1k^{1/2} + \frac{1}{3}\gamma_1k^{1/3}}\right\} \\
                &\leq\exp\left\{-k^{1/6}\frac{3\gamma_1}{14}\right\}\\
                &\leq e^{-k^{1/6}C(\gamma_1)}
            \end{align*}
        Thus in turn, if $|d_1 - \ol{d}| \leq \gamma_1 k^{1/3}$, 
            \begin{align*}
                d_1 &\geq \ol{d} - \gamma_1 k^{1/3}\\
                &\geq (\gamma_1+\gamma_2)k^{1/2} - \gamma_1k^{-1/2} - \gamma_1 k^{1/3}\\
                &\geq \frac{\gamma_1+\gamma_2}{2}k^{1/2}
            \end{align*}
        for $k$ sufficiently large. The claim follows by the union bound.
    \end{proof}


    \begin{proof}[Proof of \cref{lem:min-degree-sharp}]
        Using the same setup as in the proof of \cref{lem:min-degree-dense}, we begin by fixing $i$, which without loss of generality we take to be $i=1$, and observing that
            \begin{align*}
                d_1 = \sum_{j=2}^{k}\mathbf{1}_{j\sim i} + \sum_{j=k+1}^{2k}\mathbf{1}_{j\sim i}.
            \end{align*}
        Note that for $2\leq j \leq k$, we have $\mathbf{1}_{j\sim i}\sim\mathsf{Bernoulli}(\gamma_1 \log{k}/k)$, and for $k+1\leq j \leq 2k$, we have $\mathbf{1}_{j\sim i}\sim\mathsf{Bernoulli}(\gamma_2 \log{k}/k)$. Denote $\ep{d_1} = \ol{d}$. By the multiplicative Chernoff bound (see, e.g.,~\cite[Theorem 4.5]{mitzenmacher2017probability}), we have that for each $0<\delta<1$, it holds
            \begin{align*}
                \pr{d_1 \leq (1-\delta)\ol{d}} \leq e^{-\delta^2\ol{d}/2}.
            \end{align*}
        We note that
            \begin{align*}
                e^{-\delta^2\ol{d}/2}&=\exp\left\{-\frac{\delta^2}{2}\left((\gamma_1+\gamma_2)\log{k} + \gamma_1\log{k}/k\right)\right\}\\
                &\leq \exp\left\{-\delta^2\log{k} \frac{\gamma_1+\gamma_2}{2}\right\}
            \end{align*}
        With $\gamma_1>\gamma_2>1$, we have $\frac{2}{\gamma_1+\gamma_2}<1$, $\delta = \left(\frac{2(1+\theta)}{\gamma_1+\gamma_2}\right)^{1/2}$ for $\theta>0$ small enough, we have
            \begin{align*}
                \exp\left\{-\delta^2\log{k} \frac{\gamma_1+\gamma_2}{2}\right\} &\leq k^{-(1+\theta)}.
            \end{align*}
        By the union bound, it holds
            \begin{align*}
                \pr{d_{\min} \geq \theta\ol{d}}\geq 1-k^{-\theta}
            \end{align*}    
        The claim follows.
    \end{proof}

\end{document}

%% file: ssbm_picture_hist_1.tex
\begin{tikzpicture}[scale=0.65]

                \begin{axis}[
                    ytick = \empty,
                    name = plot00,
                    ybar,
                    ymin=0
                ]
                \addplot +[
                    hist={
                        bins=50,
                        data min=-0.05,
                        data max=1.35
                    },
                    fill=blue!30,
                    draw=blue
                ] table [y index=0] {ssbm_eigvals_00.csv};
                \draw [<-] (axis cs:0.1893, 2.5)-- +(0pt,10pt) node[above] {$\lambda_1\approx 0.19$};
                \end{axis}
                \begin{axis}[
                    ytick = \empty,
                    name = plot01,at={(plot00.east)},anchor=west,xshift=1cm,
                    ybar,
                    ymin=0
                ]
                \addplot +[
                    hist={
                        bins=50,
                        data min=-0.05,
                        data max=1.35
                    },
                    fill=blue!30,
                    draw=blue
                ] table [y index=0] {ssbm_eigvals_01.csv};
                \draw [<-] (axis cs:0.3679, 2.5)-- +(0pt,10pt) node[above] {$\lambda_1\approx 0.37$};
                \end{axis}
                \begin{axis}[
                    ytick = \empty,
                    name = plot02,at={(plot01.east)},anchor=west,xshift=1cm,
                    ybar,
                    ymin=0
                ]
                \addplot +[
                    hist={
                        bins=50,
                        data min=-0.05,
                        data max=1.35
                    },
                    fill=blue!30,
                    draw=blue
                ] table [y index=0] {ssbm_eigvals_02.csv};
                \draw [<-] (axis cs:0.5692, 2.5)-- +(0pt,10pt) node[above] {$\lambda_1\approx 0.57$};
                \end{axis}\end{tikzpicture}